\documentclass[11pt]{article}
\usepackage{amsmath}
\usepackage{amsthm}
\usepackage{amssymb}
\usepackage{color}
\usepackage{graphicx,psfrag,epsf}
\usepackage{enumerate}
\usepackage{natbib}
\usepackage{authblk}
\usepackage{parskip}
\usepackage{url} 

\usepackage{amsfonts,bm}
\usepackage{xcolor}
\usepackage{booktabs}
\usepackage{algorithm}
\usepackage{algorithmic}
\usepackage{subcaption}
\usepackage{caption}
\usepackage{longtable}
\usepackage{float}
\usepackage{xr-hyper}
\usepackage{hyperref}
\usepackage{threeparttable}
\usepackage{multirow}
\usepackage{comment}

\newcommand{\blind}{0}

\newtheorem{assumption}{Assumption}

\newtheorem{remark}{Remark}

\newcommand{\V}[1]{\boldsymbol{#1}}
\newcommand{\E}{\mathbb{E}}

\newcommand{\tr}{\mathrm{tr}}

\newtheorem{theorem}{Theorem}

\begin{document}

\def\spacingset#1{\renewcommand{\baselinestretch}%
{#1}\small\normalsize} \spacingset{1}

\if0\blind
{
	{\title{\bf Deep Jump Gaussian Processes for Surrogate Modeling of High-Dimensional Piecewise Continuous Functions}}

\author{Yang Xu$^{1}$, Chiwoo Park$^{1}$ \\
\small{$^{1}$ Department of Industrial and Systems Engineering, University of Washington, Seattle WA 98195, USA.}
}
\date{}
\maketitle
} \fi
\if1\blind
{
	\title{\bf Deep Jump Gaussian Processes for Surrogate Modeling of High-Dimensional Piecewise Continuous Functions}
	\author{}
	\date{}
	\maketitle
} \fi

\vspace{-30pt}
\begin{abstract}
We introduce Deep Jump Gaussian Processes (DJGP), a novel method for surrogate modeling of a piecewise continuous function on a high-dimensional domain. DJGP addresses the limitations of conventional Jump Gaussian Processes (JGP) in high-dimensional input spaces by integrating {target-specific}, locally linear projections with JGP modeling. These projections employ target-specific matrices to capture
local subspace structures, making them well-suited to the inherently localized modeling
behavior of JGPs, a variant of local Gaussian processes. To control model complexity,
we place a Gaussian {process (GP)} prior on the  projection matrices, allowing them to evolve
smoothly across the input space. The projected inputs are then modeled with a JGP
to capture piecewise continuous relationships with the response. This yields a distinctive {two-layer GP/JGP architecture}. We develop a scalable variational inference
algorithm to jointly learn the projection matrices and JGP hyperparameters.
We derive a finite-sample risk decomposition that separates approximation,
gating, projection, and local-linearization errors from the remaining latent-space
GP regression error.
Experiments on
synthetic and benchmark datasets demonstrate that DJGP achieves the strongest overall accuracy--calibration trade-off{, obtaining the best average rank on all three metrics}.

\end{abstract}

\noindent%

\spacingset{1.5} 

\section{Introduction}

This paper addresses surrogate modeling of piecewise continuous system responses in high-dimensional input spaces. In many engineering and scientific domains, system responses can exhibit abrupt jumps or sharp transitions under small input perturbations. For instance, in geostatistics, subsurface rock properties such as porosity and permeability can change dramatically at sedimentary interfaces, naturally giving rise to piecewise continuous behavior \citep{chiles2012geostatistics}. In materials science, first-order phase transitions (e.g., the ferromagnetic–paramagnetic shift at the Curie point) induce discontinuous changes in properties like magnetization and density \citep{park2022sequential}. In econometrics, regression discontinuity designs leverage sharp outcome changes at policy thresholds or eligibility cutoffs to identify causal effects \citep{kang2019errors}. In smart manufacturing systems, system performance may change abruptly as operating conditions approach capacity constraints \citep{Park23102025}. Developing surrogate models for piecewise continuous response surfaces is essential for data-driven understanding and uncertainty quantification of such systems.

While Gaussian processes (GPs) offer flexible Bayesian nonparametric surrogate models with uncertainty quantification, they typically rely on stationary kernels such as the squared exponential, which induce strong correlations between nearby inputs and impose smoothness, making them poorly suited for modeling abrupt changes or discontinuities~\citep{park2022jump}. Nonstationary GP models can better adapt to these changes by adjusting their hyperparameters locally to capture varying covariance structures \citep{sampson1992nonparametric,sauer2023non}. Representative approaches include heteroscedastic GPs \citep{kersting2007most,quadrianto2009kernel} and latent GPs that model kernel parameters such as variance or lengthscale \citep{paciorek2003nonstationary,tolvanen2014expectation,heinonen2016non}. A more flexible alternative is Deep Gaussian Processes (DGPs) \citep{lawrence2007hierarchical,damianou2013deep}, which stack multiple GP layers to warp inputs into nonlinear feature spaces and map them to responses. This hierarchical structure enables DGPs to capture complex nonstationary patterns that shallow GPs cannot represent, but this flexibility comes {at the cost of intractable inference}. Considerable effort has gone into scalable inference for DGPs, including Vecchia approximations \citep{Sauer03072023}, variational frameworks \citep{titsias2009variational,hensman2013gaussian,damianou2011variational,damianou2015deep}, and sampling-based methods \citep{havasi2018inference}. Hybrid models that combine neural network layers with GP layers further improve flexibility and scalability \citep{dai2015variational,wilson2016deep,wilson2016stochastic,lee2017deep}.
Despite these advances, most nonstationary GP models—including DGP-based variants—remain fundamentally smooth and tend to blur discontinuities. Additionally, DGPs are often data-hungry and require large training sets.

An approach directly suited for modeling piecewise continuous surrogates is the partitioned GP, which divides the input space into regions and fits an independent GP within each region. When the partitions align well with discontinuities, partitioned GPs can effectively represent piecewise continuous surrogates. {However, allowing highly flexible partitions requires searching over complex boundary structures and fitting multiple local GPs, which increases both model complexity and computational cost.} To control model complexity and computational cost, existing methods typically constrain how the space is partitioned. Common approaches include tessellation-based methods (e.g., Voronoi diagrams) \citep{kim2005analyzing,pope2021gaussian,luo2021bayesian} and treed partitioning \citep{gramacy2008bayesian,konomi2014adaptive,taddy2011dynamic}. These approaches improve scalability but rely on simplistic domain partitioning, making them less effective for complex or nonlinear boundaries.

A recent and more flexible approach is the Jump Gaussian Process (JGP)~\citep{park2022jump}. Rather than explicitly modeling a global partition of the input space, JGP constructs local approximations of the partition boundaries. If the boundary is sufficiently smooth, it can be locally approximated by a linear or low-order polynomial function. At each test location, JGP fits both the local polynomial boundary and the local GP parameters using data within a small neighborhood of the test location. By leveraging these local approximations, JGP can represent piecewise continuous surrogates with highly complex regional boundaries. However, JGP faces challenges in high-dimensional settings. As input dimensionality increases, data sparsity grows, requiring larger neighborhoods to obtain sufficient local training data. This leads to coarser approximations and ultimately limits JGP’s ability to capture fine-grained local structures in high-dimensional spaces.

The limitations of JGP motivate us to investigate dimensionality reduction for JGP. {A natural approach to mitigating the dimensionality problem is to apply dimensionality reduction before GP modeling. Examples include linear methods such as principal component analysis (PCA)~\citep{abdi2010principal}, as well as nonlinear methods such as kernel PCA~\citep{scholkopf1997kernel}, Isomap~\citep{tenenbaum2000global}, locally linear embedding~\citep{Roweis2000}, autoencoders~\citep{wang2016auto}, and t-SNE~\citep{maaten2008visualizing}.} However, these methods are unsupervised, meaning that they rely only on input data and ignore correlations between transformed features and the response variable. A more informative alternative is to use supervised methods such as sliced inverse regression (SIR)~\citep{li1991sliced}, supervised autoencoders~\citep{makhzani2015winner,le2018supervised}, and conditional variational autoencoders (CVAEs)~\citep{sohn2015learning,kingma2014semi}. Nevertheless, these supervised dimension-reduction techniques are still not {explicitly tailored to} the downstream GP modeling task.

Dimensionality reduction can be optimized directly for a target GP modeling task. For instance, the Mahalanobis Gaussian Process {(MGP)~\citep{aueb2013variational}} learns a linear projection of the inputs to a a low-dimensional space, with the linear projection matrix optimized jointly with the GP model parameters. The Gaussian Process Latent Variable Model {(GP-LVM)~ \citep{titsias2010bayesian}} generalizes the linear projection with a nonlinear projection represented by a GP model, resulting in a two-layer GP model with the first layer for non-linear feature mapping and the second layer for mapping to the response variable. The Deep Mahalanobis Gaussian Process {(DMGP)~\citep{dedeep}} extends MGP with a similar two-layer design. It introduces a distinct linear projection matrix for each input location, with GP priors enforcing smooth variation of these matrices across the input space. The projected features are passed to another GP layer to model the response.

Nevertheless, existing built-in dimensionality reduction methods are not tailored for JGP. Their integration is nontrivial due to a fundamental modeling difference: conventional GP models follow an inductive learning paradigm, whereas JGP operates in a transductive setting—fitting a local model at each test location. The goal of this paper is to develop a built-in dimensionality reduction approach specifically designed for JGP, yielding a piecewise continuous surrogate model that remains effective in high-dimensional input spaces.

Our approach employs locally linear projections to map high-dimensional inputs to low-dimensional latent features. For each test location, we define a location-specific linear projection matrix, with spatial correlations across these matrices enforced via a GP prior. The resulting local latent features are then mapped to the response using a JGP model, yielding a unique two-layer GP/JGP architecture, which is referred to as Deep Jump Gaussian Processes (DJGP).

The remainder of this paper is organized as follows. Section~\ref{sec:background} reviews background on stationary GPs, jump GPs, and Mahalanobis GPs. {Section~\ref{sec:lmjgp} presents the proposed DJGP model, including its local
approximation, variational inference, and prediction procedure.} Section~\ref{sec:theorem}
presents a finite-sample prediction-risk decomposition. {Section~\ref{sec:simulations} reports the primary synthetic benchmark results and summarizes additional robustness and sensitivity analyses, with the complete scaling, misspecification, and tuning results provided in Sections~D.1--D.3 of the supplement.} Section~\ref{sec:real} reports results on real-world datasets. Finally, Section~\ref{sec:conclusion} concludes this paper.

\section{Review}\label{sec:background}
Here we provide a brief technical review of the key model components: stationary GPs, jump GPs, and Mahalanobis GPs.

\subsection{Stationary Gaussian Process Surrogates} \label{sec:gp}

Consider an unknown function $f: \mathcal{X} \to \mathbb{R}$ that maps an input $\V{x} \in \mathcal{X}$ to a real response $y$, where $\mathcal{X} \subset \mathbb{R}^D$ denotes the input space. Given noisy evaluations $y_i \,\overset{\mathrm{i.i.d.}}{\sim}\, \mathcal{N}\bigl(f(\V{x}_i),\, {\sigma_{\varepsilon}^{2}}\bigr)$ for $i = 1, \dots, N$, we can build a stationary GP surrogate for $f$ by assigning a stationary Gaussian process prior with constant mean function $\mu$ and covariance kernel $c(\cdot, \cdot)$:
$
f(x) \sim \mathcal{GP}\bigl(\mu, c(\cdot, \cdot;\V{\theta})\bigr).
$
The covariance kernel is a positive definite function $c: \mathcal{X} \times \mathcal{X} \rightarrow \mathbb{R}$ parameterized by hyperparameters $\V{\theta}$. Under stationarity, the covariance depends only on the relative displacement between inputs. A widely used stationary kernel is the squared exponential (SE):
\begin{equation} \label{eq:cov_se}
{c_{\mathrm{SE}}}(\V{x}_i, \V{x}_j; \sigma_f, \ell_1, \ldots, \ell_D) =
\sigma_f^2 \exp\left(-\tfrac{1}{2} \sum_{m=1}^D \frac{(x_{im} - x_{jm})^2}{\ell_m^2} \right),
\end{equation}
where $x_{im}$ denotes the $m$th dimension of $\V{x}_i$. For the SE kernel, $\V{\theta}=(\sigma_f,\ell_1,\ldots,\ell_D)$. The Matérn class provides greater flexibility by controlling the smoothness of the function \citep{wendland2004scattered}.

Under this prior, the vector of the observed outputs $\V{y}_N = [y_1, \dots, y_N]^\top$ follows a multivariate normal distribution,
$
\V{y}_N \sim \mathcal{N}\bigl(\mu \mathbf{1}_N,\, {\sigma_{\varepsilon}^{2}} \V{I}_N + \V{C}_N \bigr),
$
where $\V{C}_N$ is an $N\times N$ matrix with its $(i,j)$th element equal to $c(\V{x}_i, \V{x}_j;\V{\theta})$, $\V{1}_N$ is an $N$-dimensional column vector of ones, and $\V{I}_N$ is an $N$-dimensional identity matrix.

The hyperparameters $\V{\theta}$, mean parameter $\mu$, and noise variance ${\sigma_{\varepsilon}^{2}}$ {can be learned} by maximizing the log marginal likelihood:
\[
L(\V{\theta}, \mu, {\sigma_{\varepsilon}^{2}})
= -\tfrac{1}{2} (\V{y}_N - \mu \mathbf{1}_N)^\top \bigl[{\sigma_{\varepsilon}^{2}} \V{I}_N + \V{C}_N\bigr]^{-1} (\V{y}_N - \mu \mathbf{1}_N)
  - \tfrac{1}{2} \log \bigl| {\sigma_{\varepsilon}^{2}} \V{I}_N + \V{C}_N \bigr|
  - \tfrac{N}{2} \log(2\pi),
\]
using either gradient-based optimization or expectation-maximization (EM)-style iterative schemes, depending on the model setting~\citep{santner2003design, gramacy2020surrogates, titsias2009variational}. We use hats, as in $\hat{\mu}$, ${\hat{\V{\theta}}}$, and ${\hat{\sigma}_{\varepsilon}^{2}}$, to denote estimated parameters.

Given the parameter estimates, we can derive the {posterior} predictive distribution of $f$ at a test input $\V{x}_* \in \mathcal{X}$. The joint distribution of $\V{y}_N$ and the latent function value $f(\V{x}_*)$ is multivariate normal (MVN). Applying the standard Gaussian conditioning formula gives the posterior predictive distribution of {$f(\V{x}_*)$}, which is Gaussian with the following {posterior} predictive mean and variance:
\[
\begin{aligned}
\mathbb{E}[{f(\V{x}_*)}] &=\hat{\mu} + {\V{c}_N}^\top \bigl[{\hat{\sigma}_{\varepsilon}^{2}} \V{I}_N + \V{C}_N \bigr]^{-1} (\V{y}_N - \hat{\mu}\V{1}_N), \\
\mathrm{Var}({f(\V{x}_*)}) &= c(\V{x}_*, \V{x}_*; {\hat{\V{\theta}}}) - \V{c}_N^\top \bigl[{\hat{\sigma}_{\varepsilon}^{2}} \V{I}_N + \V{C}_N\bigr]^{-1} \V{c}_N,
\end{aligned}
\]
where $\V{c}_N = [c(\V{x}_i, \V{x}_*;  {\hat{\V{\theta}}}): i = 1,\dots,N]$ is an $N \times 1$ vector of {the covariance values between the training inputs and the test input}.

Stationary GPs offer flexible modeling, closed-form prediction, and principled uncertainty quantification. Despite these advantages, GPs scale poorly with data size, requiring \(\mathcal{O}(N^3)\) time and \(\mathcal{O}(N^2)\) memory, which limits their applicability to moderately sized datasets. Moreover, in many practical settings, the underlying regression function is piecewise continuous and exhibits abrupt changes across unknown boundaries—behavior that stationary GPs are ill-equipped to model. Standard kernels impose global smoothness assumptions, leading to spurious correlations across discontinuities and biased estimates near regime shifts. Since stationary kernels depend solely on pairwise distances, they struggle to capture abrupt changes or heteroscedastic patterns. The Jump Gaussian Process \citep{park2022jump} addresses these limitations.

\subsection{Jump Gaussian Processes} \label{sec:jgp}
JGP is best understood through the lens of local GP modeling  {(LAGP)~\citep{gramacy2015local}}.  For each test location $\V{x}_* \in \mathcal{X}$, a small subset of nearby training data is selected, $\mathcal{D}_n^{(*)} = \{(\V{x}_{i}^{(*)}, y_{i}^{(*)}) \}_{i=1}^n$, and a conventional stationary GP model is fitted to this local data. This approach is computationally efficient—$\mathcal{O}(n^3)$ versus $\mathcal{O}(N^3)$ when $n \ll N$—and can be massively parallelized across many test points \citep{gramacy2014massively}.  A key limitation of LAGP in estimating piecewise continuous surrogates is that local neighborhoods
$\mathcal{D}_n^{(*)}$ may overlap partially or fully with {discontinuities}. In such cases, LAGP can yield biased predictions \citep{park2022jump} because the local data may mix training examples drawn from regions of the input space separated by abrupt regime shifts.

JGP addresses this issue by explicitly dividing the local data into two groups based on regime membership: data in the same regime as the test input $\V{x}_*$ and the remainder. To accomplish this, JGP introduces a latent binary random variable \(v_{i}^{(*)}\in\{0,1\}\) indicating whether a training input $\V{x}_{i}^{(*)}$ belongs to the same regime as \(\V{x}_*\) ($v_{i}^{(*)} = 1$) or not ($v_{i}^{(*)}=0$). Conditional on $v_{i}^{(*)}$ values, $i=1,\dots,n$, the local data $\mathcal{D}_n^{(*)}$ is partitioned into two groups: $\mathcal{D}_{*} = \{i \in \{1,\ldots, n\}: {v_i^{(*)}=1} \}$ and $\mathcal{D}_o = \{1,\ldots, n\} \backslash \mathcal{D}_*$.

Only \(\mathcal{D}_*\) contributes to predicting $f$ at $\V{x}_*$, while data in \(\mathcal{D}_o\) are down-weighted via a uniform “outlier” likelihood. The full specification is completed by modeling $\mathcal{D}_*$ with a stationary GP [Section \ref{sec:gp}], $\mathcal{D}_o$ with a flat ``outlier'' likelihood $p(y_{i}^{(*)} \mid v_{i}^{(*)} = 0) {= 1/u}$ for some constant $u>0$, and assigning a prior to the latent variable $v_{i}^{(*)}$, via a sigmoid function $\pi$ applied to a partitioning function $h(\V{x}; {\V{\nu}^{(*)}})$,
\begin{equation} \label{eq:latent}
	p(v_{i}^{(*)} = 1|\V{x}_{i}^{(*)}, {\V{\nu}^{(*)}}) = \pi(h(\V{x}_{i}^{(*)}; {\V{\nu}^{(*)}})),
\end{equation}
where ${\V{\nu}^{(*)}}$ is another hyperparameter{, again local to $\V{x}_*$}.
{The choice of the parametric partitioning function $h$ determines the
boundary separating $\mathcal{D}_o$ and $\mathcal{D}_*$. The linear gate
should be interpreted as a local, rather than global, boundary assumption.
When the jump boundary is sufficiently smooth within the neighborhood of
$\V{x}_*$, its first-order Taylor approximation is affine. This motivates
the local linear form
\[
h(\V{x};\V{\nu}^{(*)})=\bigl(\V{\nu}^{(*)}\bigr)^{\!\top}[1,\V{x}],
\]
even when the global boundary is nonlinear. A quadratic or other low-order
polynomial gate may reduce approximation bias for strongly curved local
boundaries, but it introduces additional parameters and may be less stable
given the small local sample size.}

Specifically, for $\V{v}^{(*)} = (v_{i}^{(*)})_{i=1}^n$ and $\V{f}^{(*)} = (f(\V{x}_{i}^{(*)}))_{i=1}^n$,
the JGP model is summarized as follows:
\begin{equation}\label{eq:jgp_local_param}
\begin{aligned}
	p(\V{y}_{n} \mid \V{f}^{(*)}, \V{v}^{(*)}, {\sigma_{\varepsilon}^{2,(*)}}) &= \prod_{i=1}^n \mathcal{N}_1(y_{i}^{(*)}|f_{i}^{(*)}, {\sigma_{\varepsilon}^{2,(*)}})^{v_{i}^{(*)}} {\left(\tfrac{1}{u}\right)^{1-v_{i}^{(*)}}},\\
	p(\V{v}^{(*)} \mid {\V{\nu}^{(*)}}) &= \prod_{i=1}^n \pi(h(\V{x}_{i}^{(*)};{\V{\nu}^{(*)}}))^{v_{i}^{(*)}} (1-\pi(h(\V{x}_{i}^{(*)}; {\V{\nu}^{(*)}})))^{1-v_{i}^{(*)}}, \\
	p(\V{f}^{(*)} \mid {\mu^{(*)}}, {\V{\theta}^{(*)}}) &= \mathcal{N}_n(\V{f}^{(*)} \mid {\mu^{(*)}}\V{1}_n, \V{C}_{n}),
\end{aligned}
\end{equation}
where $\V{y}_{n} = (y_{i}^{(*)})_{i=1}^n$ and $\V{C}_{n}$ is an $n \times n$ matrix with $c({\V{x}}_{i}^{(*)}, {\V{x}}_{j}^{(*)}; {\V{\theta}^{(*)}})$ as its $(i,j)$th element and {$\mathcal N_d(\,\cdot\mid\boldsymbol{\mu},\boldsymbol{\Sigma})$
denotes the $d$-variate Gaussian density.}
These local quantities are estimated separately for every test location $\V{x}_*$ from its own neighborhood $\mathcal{D}_n^{(*)}$.
Parameters and latent indicators are learned via an EM‐style algorithm, e.g., a variational EM variant (JGP‐VEM) approximates the joint posterior over \(\{\V{v}^{(*)},\V{f}^{(*)}\}\), yielding similar predictive equations with uncertainty propagation.

Let $\hat{v}_i^{(*)}$ represent the maximum a posteriori (MAP) estimate of $v_i^{(*)}$ upon convergence of the EM algorithm and let $\hat{\mathcal D}_{n}^{(*)}=\{i:\,\hat{v}_i^{(*)}=1\}$ denote the estimated in‐regime subset and $\hat{\mathcal D}_o^{(*)}=\{1,\ldots,n\}\setminus\hat{\mathcal D}_{n}^{(*)}$ the out‐of‐regime subset. Let $\V{y}_* = (y_i^{(*)}, i \in \hat{\mathcal D}_{n}^{(*)})$ and {$\widehat n_*$} denote the number of {}elements in $\hat{\mathcal D}_{n}^{(*)}$. The posterior predictive mean and variance for $f(\V x_*)$ are
\begin{equation}
\label{eq:JGP_post_pred}
\begin{aligned}
\mu_* &= {\hat{\mu}^{(*)}} + {\V{c}_{\widehat n_*,*}}^{\top}\!\bigl({\hat{\sigma}_{\varepsilon}^{2,(*)}}{\V I_{\widehat n_*}} + {\V{C}_{\widehat n_*}^{(*)}}\bigr)^{-1}
\bigl(\V y_*-{\hat{\mu}^{(*)}}\V 1\bigr),\\
\sigma_*^{2} &= c(\V{x}_*, \V{x}_*; \hat{\V{\theta}}^{(*)})
-{\V{c}_{\widehat n_*,*}}^{\top}\!\bigl({\hat{\sigma}_{\varepsilon}^{2,(*)}}{\V I_{\widehat n_*}} + {\V{C}_{\widehat n_*}^{(*)}}\bigr)^{-1}{\V{c}_{\widehat n_*,*}},
\end{aligned}
\end{equation}
where ${\V{c}_{\widehat n_*,*}} = (c(\V{x}^{(*)}_i, \V{x}_*; \hat{\V{\theta}}^{(*)}))_{i \in \hat{\mathcal D}_{n}^{(*)}}$ is a ${\widehat n_* \times 1}$ column vector of {the covariance values between the retained local inputs and the test input $\V x_*$}, and ${\V{C}_{\widehat n_*}^{(*)}} = (c(\V{x}^{(*)}_i, \V{x}^{(*)}_j; \hat{\V{\theta}}^{(*)}))_{i, j \in \hat{\mathcal D}_{n}^{(*)}}$ is a ${\widehat n_* \times \widehat n_*}$ {square matrix of covariances evaluated at all pairs of retained local inputs}. Here, ${\hat{\sigma}_{\varepsilon}^{2,(*)}}$,  $\hat{\V{\theta}}^{(*)}$ and ${\hat{\mu}^{(*)}}$ represent the maximum-likelihood estimates of ${\sigma_{\varepsilon}^{2,(*)}}$,  $\V{\theta}^{(*)}$ and ${\mu^{(*)}}$, respectively.

When the input dimension \(D\) is large, several challenges arise for JGP modeling.  First, the number of hyperparameters grows quickly: a linear partition function \(h(\V{x};{\V{\nu}^{(*)}})\) requires \(D+1\) parameters, while a quadratic function demands on the order of \(D^2+1\) parameters, quickly overwhelming the modest size of local neighborhoods.  Second, there is a fundamental trade-off between bias and variance: enlarging the neighborhood yields more data for stable estimation of \({\V{\nu}^{(*)}}\), but weakens the fidelity of the local Taylor approximation to complex boundaries; conversely, restricting to a small neighborhood preserves locality but risks overfitting due to limited data. Finally, the curse of dimensionality leads to sparse coverage in high-dimensional spaces, making it difficult to learn reliable regime boundaries without prior dimension reduction. These limitations motivate a unified framework that integrates dimensionality reduction directly into the JGP model, thereby enabling more effective modeling of high-dimensional, piecewise continuous functions.

\subsection{Mahalanobis Gaussian Processes}

Mahalanobis Gaussian Processes~\citep{aueb2013variational} extend traditional Gaussian process models by incorporating a built-in dimensionality reduction. The input vector $\V{x}$ is linearly projected to $\V{W}\V{x}$ by a linear projection matrix \(\V{W} \in \mathbb{R}^{K \times D}\). The relation of the projected features to the response is modeled using a stationary GP model with the covariance kernel defined on the projected features. For instance, the squared exponential covariance $\eqref{eq:cov_se}$ can be defined with the projected features as
\begin{align}\label{mgp}
    K_W(\V{x}_i,\V{x}_j; \V\theta) = \sigma_f^2 \exp\left(-\frac{1}{2}(\V{x}_i - \V{x}_j)^\top \V{W}^\top \V{W} (\V{x}_i - \V{x}_j)\right).
\end{align}
To enable tractable learning, the authors introduced a variational inference framework for jointly estimating $\boldsymbol{W}$ along with the remaining GP hyperparameters.

Deep Mahalanobis Gaussian Process (DMGP)~\citep{dedeep} extends MGP by introducing a nonlinear, input-dependent projection. Unlike MGPs, which employ a single global linear projection $\boldsymbol{W}$, DMGP assigns each data point $\boldsymbol{x}$ its own projection matrix $\boldsymbol{W}(\boldsymbol{x})$. This results in a non-linear projection $g(\V{x}) = \V{W}(\V{x})\V{x}$. Each entry of $\boldsymbol{W}(\boldsymbol{x})$ is modeled as a function of $\boldsymbol{x}$, governed by a stationary GP. Within each row of $\boldsymbol{W}(\boldsymbol{x})$, the elements share a common GP prior with identical kernel hyperparameters. This construction enforces equal scaling of elements within a row through the shared kernel variance parameter, thereby achieving automatic relevance determination for the associated latent dimension.

DMGP models a spatially varying projection matrix, $\boldsymbol{W}(\boldsymbol{x})$, using a multivariate Gaussian process. Directly applying this projection scheme to a local GP model such as JGP introduces several technical challenges. In particular, DMGP assigns a distinct local projection matrix to each input location. Consequently, its integration with a local GP may require estimating a separate high-dimensional projection matrix for every training/test point in a local neighborhood. This substantially increases the dimension of the latent parameter space and weakens identifiability between the projection variables and the other JGP parameters, given that JGP is fitted using a small local sample size $n$. This motivates our main contribution. {In DJGP}, we seek a locally constant approximation of $\V{W}(\V{x})$. When $\V{W}(\V{x})$ is a smooth function of $\V{x}$, the local constant approximation can be justified by the zeroth-order Taylor approximation. Specifically, $\V{W}(\V{x})$ is approximately equal to $\V{W}(\V{x}_*)$ for the local data $\mathcal{D}_n^{(*)}$ near a test location $\V{x}_*$. Under this formulation, the JGP model for $\boldsymbol{x}_*$ requires only a single projection matrix in addition to its standard parameters. This substantially reduces the number of parameters to estimate, improving feasibility while still capturing nonlinear projections through a piecewise constant structure.

\section{Deep Jump Gaussian Process}\label{sec:lmjgp}

Let $\mathcal{X}$ denote a domain of a function in $\mathbb{R}^D$. We consider a problem of estimating an unknown surrogate function which relates inputs $\V{x} \in \mathcal{X}$ to a real response variable. We assume the existence of a nonlinear sufficient dimension reduction (for the unknown surrogate relation), which reduces the $D$-dimensional feature in $\mathcal{X}$ to a lower-dimensional feature in $\mathcal{Z} \subseteq \mathbb{R}^K$ via a continuously differentiable mapping,
\[
g: \mathcal{X} \;\longrightarrow\; \mathcal{Z}\subseteq\mathbb{R}^K,\qquad K\ll D,
\]
so that the response variable depends on $\boldsymbol{x}$ only through its reduced representation $\boldsymbol{z} = g(\boldsymbol{x})$. Therefore, we can introduce a reduced surrogate model $f$ to relate $\V{z}$ to the response variable. We assume $f$ is piecewise continuous in $\V{z}$, so the composition function $f\circ g$ is also piecewise continuous in $\V{x}$, given the assumed continuity of $g$. Specifically, there exists an (unknown) integer $M$ and an unknown partition of $\mathcal{Z}$ into disjoint regions $\{\mathcal{Z}_m\}_{m=1}^M$ such that
\begin{equation} \label{eq:mixture}
f(\boldsymbol{z}) \;=\; \sum_{m=1}^M f_m(\boldsymbol{z})\,1_{\mathcal{Z}_m}(\boldsymbol{z}),
\end{equation}
where each local function $f_m$ is a continuous function with its uncertainty modeled as a stationary Gaussian process with constant mean $\mu_m \in \mathbb{R}$ and a stationary covariance function $c_m(\cdot,\cdot)$. We assume mutual independence across regions:
\begin{equation} \label{eq:indep}
	\mbox{Independence:} \qquad f_m \mbox{ is independent of } f_{\ell} \mbox{ for } m \neq \ell.
\end{equation}
This implies zero correlation between function values belonging to different regions.

For simplicity, we restrict $c_m$ to a parametric family ${\{c(\cdot,\cdot;\V{\theta})\}}$, though the framework extends naturally to more general covariance functions. Let $\V{\theta}_{m}$ denote the region-specific covariance parameter so that $c_m(\cdot,\cdot) = c(\cdot,\cdot;\V{\theta}_m)$. We specifically consider the scale family, $c(\V{z},\V{z}'; \V{\theta}_m) = a_m C ( b_m||\V{z} - \V{z}'||_2 ),$
where $||\V{z} - \V{z}'||_2$ is the Euclidean distance, $C(\cdot)$ is an isotropic correlation function of the distance with a unit length scale (for the squared-exponential kernel, $C(r)=\exp(-r^2/2)$ with $r=\|\V z-\V z'\|_2$), $a_m > 0$ is the variance parameter, and $b_m > 0$ is the inverse length-scale parameter.

Finally, we assume heterogeneity in {regional response distributions}:
\begingroup
\begin{equation} \label{eq:heter}
	\mbox{Heterogeneity:} \qquad (\mu_m,\V{\theta}_m) \neq (\mu_{\ell},\V{\theta}_{\ell}) \quad \text{for adjacent } m \neq \ell.
\end{equation}
That is, adjacent regions differ in either their mean or their covariance structure. If two adjacent regions share an identical response distribution but exhibit no realized jump, their common boundary is unidentifiable; in that degenerate case merging the two regions does not change predictions.
\endgroup

We aim to predict the surrogate response $f\circ g(\V{x})$ at $J$ test locations $\{\V{x}^{(j)} \in \mathcal{X}, j = 1,\ldots, J\}$, given $N$ noisy observations from the underlying model. Each observation at $\boldsymbol{x}_i$ is given as
\begin{equation} \label{model:noise}
	y_i = f\big(g(\boldsymbol{x}_i)\big) + \epsilon_i,
	\qquad i = 1,\ldots, N,
\end{equation}
where the noises $\epsilon_i \sim \mathcal{N}(0, {\sigma_{\varepsilon}^{2}}(g(\V{x}_i)))$ are independent. We denote the total training dataset $\mathcal{D}_{\V X}=(\V X,\V y)=\{(\V{x}_i, y_i), i = 1,\ldots, N\}$.  The noise variance is assumed to change smoothly in the projected input $g(\V{x}_i)$ and is thus also smooth in the original input $\V{x}_i$, so the variance is approximately constant around a small neighborhood of the projected input $g(\V{x}_i)$.

\subsection{Local Approximation}
Modeling and estimating the complex functions $g$ and $f$ explicitly together with the unknown partition $\{\mathcal{Z}_m\}_{m=1}^M$ is challenging. Following the JGP framework, we instead seek a local approximation. For each test location $\V{x}^{(j)}_*$, we first select a small subset of nearby training data—for example, the $n$ nearest neighbors of $\V{x}^{(j)}_*$ or a subset chosen by an existing local selection criterion~\citep{gramacy2015local}. We denote this local dataset by
\begin{equation}
	\mathcal{D}_n^{(j)} = \{(\V{x}_{i}^{(j)}, y_i^{(j)}): i = 1, \ldots, n \},
\end{equation}
where $\V{x}_{i}^{(j)}$ and $y_i^{(j)}$ represent the input vector and corresponding response of the $i$th local data. For brevity, we introduce $\V X^{(j)} = \{\V{x}_{i}^{(j)}, i = 1,\ldots, n\}$ and $\V y^{(j)} = \{y_{i}^{(j)}, i = 1,\ldots, n\}$.

Since we assume the map $g$ is smooth and continuously differentiable, we apply a first-order Taylor approximation to $g$ in a neighborhood of $\V{x}^{(j)}_*$. For each local training point $(\V{x}_{i}^{(j)}, y_i^{(j)})$ within the neighborhood,
\begin{equation}
	g(\V{x}_{i}^{(j)}) \approx g(\V{x}_{*}^{(j)}) + \V{W}_j (\V{x}_{i}^{(j)} -\V{x}_*^{(j)}).
\end{equation}
The constant terms in the approximation do not affect the downstream GP modeling, so we omit the constant terms and define a local projection by $g(\V{x}_{i}^{(j)}) \approx \V{W}_j \V{x}_{i}^{(j)}$.

The projection matrix $\V{W}_j$ defines the direction of the local projection to the latent feature space of intrinsic dimension $K$. Because $K$ is unknown in practice, we need to estimate the latent dimension. We consider this as a model selection problem, which is later addressed by a cross-validation choice of the latent dimension. Let $Q$ denote the chosen latent dimension. We fit a working projection $\V{W}_j\in\mathbb{R}^{Q\times D}$. To impose statistical correlation and encourage smooth variation of the projections over $\mathcal{X}$, we place a stationary Gaussian process prior on the collection of local projection matrices $\mathbf{W} = \{\mathbf{W}_j\}_{j=1}^J$. Specifically, let $w_{kd}^{(j)}$ denote the $(k,d)$-th entry of $\mathbf{W}_j$, and define the vector \(\V{w}_{kd} = [w_{kd}^{(1)}, \dots, w_{kd}^{(J)}]\), which includes the \((k,d)\)th entries across all local projection matrices. We then model $\boldsymbol{w}_{kd}$ as a Gaussian process with zero mean and the isotropic covariance function given by
\begin{equation*}
	c_{iso}(\boldsymbol{x}^{(j)}_*, \boldsymbol{x}^{(j')}_*; s, \ell_{w,k}) = s^2 \exp\!\left(-\frac{\|\boldsymbol{x}^{(j)}_* - \boldsymbol{x}^{(j')}_*\|^2}{2\ell_{w,k}^2}\right).
\end{equation*}
The squared exponential covariance models the correlation between local projection matrices, $\boldsymbol{W}_j$ and $\boldsymbol{W}_{j'}$, as a function of the distance between their corresponding test locations, $\boldsymbol{x}^{(j)}_*$ and $\boldsymbol{x}^{(j')}_*$. All entries share a common variance $s^2$, while elements within each row share a row-specific length-scale $\ell_{w,k}$, enforcing uniform scaling and enabling automatic relevance determination for each latent dimension. The joint prior is given by
\begin{equation} \label{eq:global_prior}
	p(\mathbf{W}|\Theta_W) =
	 \prod_{k=1}^{Q} \prod_{d=1}^D \mathcal{N}\bigl(\V{w}_{kd}| \V{0}_J, \mathbf{C}_{w}^{(k)}\bigr),
\end{equation}
Here $\Theta_W=(s,\ell_{w,1},\ldots,\ell_{w,Q})$ denotes the global
projection-GP hyperparameter block.  The vector $\V{0}_J$ contains $J$ zeros,
and $\mathbf{C}_{w}^{(k)}$ is the $J\times J$ matrix whose $(j,j')$ entry is
$c_{iso}(\boldsymbol{x}^{(j)}_*,\boldsymbol{x}^{(j')}_*;s,\ell_{w,k})$.

Conditioned on the local projection $\V{W}_j$, the projected local dataset is defined as
\begin{equation}
	\mathcal{D}^{(j)}_{\V{W}_j, n} = \{(\V{z}_{i}^{(j)}, y_i^{(j)}): i = 1, \ldots, n, \V{z}_{i}^{(j)} =  \V{W}_j \V{x}_{i}^{(j)}\}.
\end{equation}
By the mixture proposition in \eqref{eq:mixture}, these local data may originate from different regions, in which case the input--response relationship cannot be captured by a single Gaussian process. We follow JGP to model the mixture data. Specifically, in the \(j\)th local region, we introduce binary latent variables \(v_i^{(j)} \in \{0, 1\}\) to indicate whether the projected training input \(\boldsymbol{z}_i^{(j)}\) belongs to the same region as the projected test point \(\V{W}_j \boldsymbol{x}_*^{(j)}\): $v_i^{(j)}=1$ if it does and $v_i^{(j)}=0$ otherwise. Based on the indicator values, the local data $\mathcal{D}^{(j)}_{\V{W}_j, n}$ can be partitioned into two groups: $\mathcal{D}^{(j,1)}_{\V{W}_j, n} = \{ i \in \{1,\ldots, n\}:v_i^{(j)}=1 \}$ and the remainder $\mathcal{D}^{(j,0)}_{\V{W}_j, n} = \{1,\ldots, n\} \backslash \mathcal{D}^{(j,1)}_{\V{W}_j, n}$. The first group belongs to the same region as the projected test location \(\V{W}_j \boldsymbol{x}_*^{(j)}\), so we use those observations to predict $f$ at the test location. The second group is independent of $f$ under assumption~\eqref{eq:indep}, so it {is excluded from prediction}.

Since we are uncertain about the indicator values, we model them as random variables and assign each indicator the JGP prior probability in~\eqref{eq:latent}:
\[
p(v_i^{(j)}=1\mid {\V{W}_j,}\V{\nu}_j) = \pi(h(\boldsymbol{z}_i^{(j)}; \V{\nu}_j)),
\]
where \(\pi(z) = 1 / (1 + e^{-z})\) is the sigmoid link function, and we use the linear decision function $h(\boldsymbol{z}_i^{(j)}; \V{\nu}_j) = \V{\nu}_j^T [1, \boldsymbol{z}_i^{(j)}]$. The logistic model divides the local data along the linear boundary $\V{\nu}_j^T [1, \boldsymbol{z}_i^{(j)}] = 0$. We orient the affine gate such that $h(\V W_j\V x_*^{(j)})>0$, so that $v_i^{(j)}=1$ denotes the side containing the target point; the equivalent target-centered parameterization is given in Appendix~A. When the boundaries of the regions $\{\mathcal{Z}_m, m = 1,{\ldots},M\}$ are sufficiently smooth, they can be {approximated locally by a hyperplane using a first-order Taylor expansion}, which justifies using a linear boundary to split the local data. {For rougher boundaries}, one can use higher-order polynomial functions.

The first group of the local data \(\{(\V{z}_{i}^{(j)}, y_i^{(j)}): v_i^{(j)} = 1\}\) and the projected test point \(\V{W}_j \boldsymbol{x}_*^{(j)}\) {belong} to the same region, denoted $m(j)$. Under model~\eqref{eq:mixture}, the input--output relationship follows a stationary Gaussian process with constant mean $\mu_{m(j)}$ and covariance function $c(\cdot,\cdot; \V\theta_{m(j)})$. The region-specific covariance function has the form
\begin{equation*}
	c(\V{z},\V{z}'; \V\theta_{m(j)}) = a_{m(j)} C( b_{m(j)}||\V{z} - \V{z}'||_2 ).
\end{equation*}
Since the local length-scale parameter is redundant with the scale of the local projection matrix $\V{W}_j$, we remove it. This leaves the regional covariance function with only the scale parameter $a_{m(j)}$:
\begin{equation*}
	c(\V{z},\V{z}'; a_{m(j)}) = a_{m(j)} C( ||\V{z} - \V{z}'||_2 ).
\end{equation*}

Based on \eqref{model:noise}, $y_i$ is a noisy realization of the Gaussian process
\begin{equation} \label{eq:ycond1}
p\bigl(y_i^{(j)}\mid f_i^{(j)},v_i^{(j)}=1,{\sigma_{\varepsilon,j}^{2}}\bigr)
=
\mathcal{N}\bigl(y_i^{(j)}\mid f_i^{(j)},{\sigma_{\varepsilon,j}^{2}}\bigr),
\end{equation}
where \({\sigma_{\varepsilon,j}^{2}} \) is a local constant approximation to ${\sigma_{\varepsilon}^{2}}(g(\V{x}))$ at $\V{x}$ around $\V{x}_*^{(j)}$. The Taylor approximation is justifiable given smoothness of ${\sigma_{\varepsilon}^{2}}(\cdot)$ and $g(\cdot)$.

The other group of the local data \(\{(\V{z}_{i}^{(j)}, y_i^{(j)}): v_i^{(j)} = 0\}\) is independent of the response variable at the projected test point \(\V{W}_j \boldsymbol{x}^{(j)}_*\). We treat them as outliers with respect to $f^{(j)}$ and assign them a uniform likelihood,
\begin{equation} \label{eq:ycond2}
	p\bigl(y_i^{(j)}\mid v_i^{(j)}=0\bigr)
	= \frac{1}{u_j}.
\end{equation}
{Here $u_j>0$ specifies a flat (uniform) outlier density for out-of-region points{; in our setting it is a constant determined locally from $\V y^{(j)}$ rather than a learned parameter, with the exact rule given in Appendix~A}.}

Collecting the local data and latent variables, define
\[
\begin{gathered}
\V{y}^{(j)}=(y_1^{(j)},\ldots,y_n^{(j)}),\qquad
\V{v}^{(j)}=(v_1^{(j)},\ldots,v_n^{(j)}),\\
\V{f}^{(j)}=(f_1^{(j)},\ldots,f_n^{(j)}).
\end{gathered}
\]
{For the $j$th local JGP, define the parameter block
\[
\Theta_j
=
\bigl(\V\nu_j,\sigma_{\varepsilon,j}^{2},\mu_{m(j)},a_{m(j)}\bigr).
\]
Together with the global projection-GP block $\Theta_W$, all model parameters are
collected as
\[
\Theta=(\Theta_W,\Theta_1,\ldots,\Theta_J).
\]}

The conditional distributions are therefore
\[
\begin{aligned}
	& p\bigl(\V{y}^{(j)}\bigr | \V{v}^{(j)},\V{f}^{(j)},\Theta_j) =\prod_{i=1}^n \biggl[
	\mathcal{N}\bigl(y_i^{(j)}\mid f_i^{(j)},{\sigma_{\varepsilon,j}^{2}}\bigr)\biggr]^{v_i^{(j)}}  \biggl[ \frac{1}{u_j} \biggr]^{1-v_i^{(j)}}, \mbox{ and } \\
	& p\bigl(\V{v}^{(j)} \bigr |{\V{W}_j,}\Theta_j) =\prod_{i=1}^n \biggl[p(v_i^{(j)}=1\mid {\V{W}_j,}\V{\nu}_j) \biggr]^{v_i^{(j)}}  \biggl[ 1- p(v_i^{(j)}=1\mid {\V{W}_j,}\V{\nu}_j) \biggr]^{1-v_i^{(j)}}.
\end{aligned}
\]
The joint distribution for the local model is
\[
\begin{aligned}
&p\bigl(\V{y}^{(j)},\V{v}^{(j)},\V{f}^{(j)}\mid \V{W}_j, \Theta_j\bigr)\\
&\quad=p\bigl(\V{y}^{(j)}\mid \V{v}^{(j)},\V{f}^{(j)},\Theta_j\bigr)
p\bigl(\V{v}^{(j)}\mid{\V{W}_j,}\Theta_j\bigr)
p\bigl(\V{f}^{(j)}\mid \V{W}_j,\Theta_j\bigr),
\end{aligned}
\]
where $p\bigl(\V{f}^{(j)}\mid \V{W}_j,\Theta_j \bigr) = \mathcal{N}(\V{f}^{(j)}\mid \mu_{m(j)} \V{1}_n, a_{m(j)}\V{C}_{nn})$, and $\V{C}_{nn}$ is an $n \times n$ matrix with $C( ||\V{z}_i^{(j)}-\V{z}_{i'}^{(j)}||_2)$ as its $(i,i')$th element.

{The DJGP training criterion is constructed from target-specific local
factors:}
\begin{equation} \label{model:full}
	p\bigl(\V{y},\V{v},\V{f}, \V{W} \bigr | \Theta) = p(\V{W}|\Theta_W) \times \prod_{j=1}^J p\bigl(\V{y}^{(j)},\V{v}^{(j)},\V{f}^{(j)}\bigr | \V{W}_j, \Theta_j),
\end{equation}
where
{
\[
\begin{gathered}
\V{y} = (\V{y}^{(1)}, \ldots, \V{y}^{(J)}),\quad
\V{v} = (\V{v}^{(1)}, \ldots, \V{v}^{(J)}),\quad
\V{f} = (\V{f}^{(1)}, \ldots, \V{f}^{(J)}).
\end{gathered}
\]
}%
\begin{sloppypar}%
Conditioned on the local projection matrix $\V{W}_j$, the conditional model $p\bigl(\V{y}^{(j)},\V{v}^{(j)},\V{f}^{(j)}\bigr | \V{W}_j, \Theta_j)$ is a local model, a JGP model specific to the local projection data $\mathcal{D}^{(j)}_{\V{W}_j, n}$. The local projection matrices are correlated through the global GP model $p(\V{W}\mid\Theta_W)$. This unique two-layer GP/JGP model is referred to as the Deep JGP (DJGP) model.
\end{sloppypar}%

\begin{remark}[Composite likelihood]\label{rem:composite}
The local neighborhoods $\mathcal{D}_n^{(j)}$ are selected independently for each target
location and may therefore overlap, so a single training observation can appear in more than
one of the $J$ local factors of~\eqref{model:full}. We therefore treat~\eqref{model:full} as a
\emph{composite likelihood}~\citep{lindsay1988composite,varin2011overview}, so the distribution
it induces is a composite posterior rather than an exact Bayesian posterior.
\end{remark}

\subsection{Variational Inference} \label{sec:inference}

The statistical inference of the model parameters $\Theta$ and the latent variables $\V{f}$, $\V{W}$, and $\V{v}$ is analytically intractable due to the nonlinear dependencies introduced by the hierarchical structure. To address this challenge, we adopt a variational inference framework, following the sparse GP methodology introduced in MGP~\citep{aueb2013variational} and DMGP~\citep{dedeep}. Specifically, we introduce two sets of inducing variables: local inducing variables for the latent functions $\V{f}$ to decouple the otherwise intractable dependencies between latent variables and hyperparameters, and global inducing variables for the projection process $\V W$ to alleviate the prohibitive computational burden associated with repeated large-scale matrix inversions, when the number of {}test locations is large.

\paragraph{Local inducing variables.}
For each test point $\V{x}^{(j)}_*$, we introduce $L_1$ local inducing inputs
$\tilde{\V{z}}^{(j)}_\ell \in \mathbb{R}^{Q}, \quad \ell = 1,\ldots,L_1,
$
with corresponding inducing outputs $r_{\ell}^{(j)}$.

{
For each target $\V x_*^{(j)}$, we select $L_1$ input locations consisting of the target input and its nearest training neighbors in the original input space. Given the initial projection $\V W_{j,0}$, these selected input locations are projected into the local coordinate system, and their projected locations are used as the local inducing inputs. The resulting local inducing-input locations are then kept fixed in the projected coordinate system throughout inference.}

The inducing outputs are defined as standardized evaluations of the latent function $f_{m(j)}$, $
r_{\ell}^{(j)} = (f_{m(j)}(\tilde{\V{z}}^{(j)}_\ell) - \mu_{m(j)})/{\sqrt{a_{m(j)}}},$
so that they are standardized with respect to the amplitude and mean hyperparameters $a_{m(j)}$ and $\mu_{m(j)}$, which improves identifiability.
Collecting them as $\V{r}^{(j)} = (r_{\ell}^{(j)})_{\ell=1}^{L_1} \in \mathbb{R}^{L_1}$, we assume the joint Gaussian prior
$
p(\V{r}^{(j)}) = \mathcal{N}\bigl(\V{r}^{(j)} \mid \V{0}_{L_1}, \V{K}_{r}^{(j)}\bigr),
$
where $\V{K}_{r}^{(j)} \in \mathbb{R}^{L_1 \times L_1}$ has entries
$[\V{K}_{r}^{(j)}]_{\ell\ell'} = C\!\left(\|\tilde{\V{z}}_{\ell}^{(j)} - \tilde{\V{z}}_{\ell'}^{(j)}\|\right)$.
Conditioned on these inducing variables, the local latent function values $\V{f}^{(j)}$ follow
{
\[
\begin{aligned}
p(\V{f}^{(j)} \mid \V{r}^{(j)}, \V{W}_j, \Theta_j)
= \mathcal{N}\Bigl(\V{f}^{(j)} \mid\;
&\mu_{m(j)}\V 1_n+\V{K}_{fr}^{(j)} (\V{K}_{r}^{(j)})^{-1}\V{r}^{(j)},\\
&a_{m(j)}\V{C}_{nn} - \V{K}_{fr}^{(j)} (\V{K}_{r}^{(j)})^{-1} (\V{K}_{fr}^{(j)})^\top\Bigr),
\end{aligned}
\]
}%
where $\V{K}_{fr}^{(j)} \in \mathbb{R}^{n \times L_1}$ with entries
$[\V{K}_{fr}^{(j)}]_{i\ell} = {\sqrt{a_{m(j)}}}\, C(\| \V{W}_j \V{x}_i^{(j)} - \tilde{\V{z}}_{\ell}^{(j)} \|)${, consistent with the unit-variance standardization $r_\ell^{(j)}=(f_{m(j)}(\tilde{\V z}_\ell^{(j)})-\mu_{m(j)})/\sqrt{a_{m(j)}}$ and $\V K_r^{(j)}$ built from the unit-scale correlation $C$}, and $\V{C}_{nn}$ denotes the kernel matrix constructed over the projected local inputs.

\paragraph{Global inducing variables.}
Similarly, for the projection process, we introduce $L_2$ global inducing inputs
$
\tilde{\V{x}}_\ell \in \mathbb{R}^D, \quad \ell=1,\ldots,L_2,
$
with inducing outputs $\V{R}_\ell \in \mathbb{R}^{Q \times D}$, aggregated as $\V{R} = (\V{R}_\ell)_{\ell=1}^{L_2} \in \mathbb{R}^{L_2 \times Q \times D}$. {When the number of test locations, $J$, is less than $L_2$, we simply use all test inputs as the global inducing inputs. Otherwise, we apply $k$-means clustering to partition the test inputs into $L_2$ clusters and use the resulting cluster centers as the global inducing inputs.
} {These global inducing inputs are held fixed throughout optimization; only their inducing-output distribution $q(\V R)$ and the model hyperparameters are learned.} The inducing outputs are assumed to be drawn from the same GP as $\V{W}_j$. Specifically, let $R_{\ell kd}$ denote the $(k,d)$-th entry of $\mathbf{R}_{\ell}$, and define the vector \(\V{R}_{:kd} = [R_{1kd}, \dots, R_{L_2kd}]\), which includes the \((k,d)\)th entries across all inducing output matrices. Then, $\V{R}_{:kd}$ is assumed to follow the same stationary GP as  $\V{w}_{kd}$. The prior distribution of $\V{R}$ is given as
\begin{equation}\label{p(R)}
    p(\V{R}|\Theta_W) = \prod_{k=1}^{Q} \prod_{d=1}^D
    \mathcal{N}\!\bigl(\V{R}_{:kd}\,\big|\, \V{0}_{L_2}, \V{K}_{\V{R}}^{(k)}\bigr),
\end{equation}
where $\V{K}_{\V{R}}^{(k)}$ is an $L_2 \times L_2$ matrix with its $(\ell, \ell')$th entry as $c_{iso}(\tilde{\V{x}}_\ell,  \tilde{\V{x}}_{\ell'}; s, \ell_{w,k})$. To reduce the computational burden when the number of test points $J$ is large, we use a sparse Gaussian process approximation. It assumes conditional independence of the elements in $\V{w}_{k,d}$ given $\V{R}_{:kd}$, yielding the conditional distribution
\[
p(\V{W}\mid \V{R}, \Theta_W)
= \prod_{k=1}^{Q} \prod_{d=1}^D  \mathcal{N}\!\left(
\V{w}_{kd}\;\middle|\;
\V{K}_{\V{WR}}^{(k)} (\V{K}_{R}^{(k)})^{-1} \V{R}_{:kd},\;
\V{\Lambda}_{\V{W}}^{(k)} - \V{K}_{\V{WR}}^{(k)} (\V{K}_{\V{R}}^{(k)})^{-1} \V{K}_{\V{WR}}^{(k)\,\top}
\right).
\]
\begin{sloppypar}%
where
$\V{\Lambda}_{\V{W}}^{(k)}$ is a $J \times J$ diagonal matrix with its $j$th diagonal element equal to $c_{iso}(\V{x}^{(j)}_*, \V{x}^{(j)}_*; s, \ell_{w,k})$, $\V{K}_{\V{WR}}^{(k)}$ is a $J \times L_2$ matrix with its $(j, \ell)$th element equal to $c_{iso}( \V{x}^{(j)}_*, \tilde{\V{x}}_\ell; s, \ell_{w,k})$.

\end{sloppypar}%

\paragraph{{Variational family and composite ELBO.}}
{We apply the variational construction to the target-wise
composite criterion in~\eqref{model:full}:}
\begin{equation} \label{model:full2}
	\begin{split}
	&p\bigl(\V{v},\V{f}, \V{W}, \V{R}, \V{r} \bigr |\V{y}, \Theta)
	 \propto p(\V{W}|\V{R}, \Theta_W) \times p(\V{R}| \Theta_W)  \\
	& \quad \times \prod_{j=1}^J p\bigl(\V{y}^{(j)}\bigr | \V{v}^{(j)},\V{f}^{(j)},\Theta_j) \times p\bigl(\V{v}^{(j)} \bigr |\Theta_j,\V W_j) \times p(\V{f}^{(j)} \mid \V{r}^{(j)}, \V{W}_j, \Theta_j) \times p(\V{r}^{(j)})
	\end{split}
\end{equation}
We approximate it with the following factorized variational distribution:
\begin{equation}
\begin{aligned}
q\bigl(\V{v},\V{f}, \V{W},\V{R}, \V{r}\bigr)
=p\bigl(\V{W}\mid \V{R},\Theta_W \bigr)\; q\bigl(\V{R}\bigr) \prod_{j=1}^J \Bigl[\, q(\V{v}^{(j)})\,p(\V{f}^{(j)}\mid \V{r}^{(j)},\V{W}_j,\Theta_j)\,q(\V{r}^{(j)}) \Bigr].
\end{aligned}
\end{equation}
where
\begin{equation}\label{eq:def of q}
\begin{aligned}
& q\bigl(\V{v}^{(j)}\bigr) = \prod_{i\in\mathcal D_n^{(j)}}q\bigl(v_i^{(j)}\bigr) \qquad\mbox{ with } q\bigl(v_i^{(j)}\bigr)=\mathrm{Bernoulli}\bigl(\rho_i^{(j)}\bigr) \\
& q\bigl(\V{r}^{(j)}\bigr)=\mathcal{N}\bigl(\boldsymbol \mu_r^{(j)},\boldsymbol \Sigma_r^{(j)}\bigr), \\
& q(\V{R}) = \prod_{\ell=1}^{L_2} \prod_{k=1}^{Q} \prod_{d=1}^D q\bigl(R_{\ell kd} \bigr) = \prod_{k,d} q(\V R_{:kd}) \qquad\mbox{ with } q\bigl(R_{\ell kd} \bigr) =\mathcal{N}\bigl(\mu_{\ell kd},\sigma_{\ell kd}^2\bigr).
\end{aligned}
\end{equation}
Here, $\V R_{:kd} := (R_{1kd},\ldots,R_{L_2kd})^\top \in \mathbb{R}^{L_2}$ denotes the slice of $\V R$ along the inducing-point index $\ell$ for fixed $(k,d)$.
Accordingly, under~\eqref{eq:def of q} we have
$q(\V R_{:kd})=\mathcal{N}(\bm\mu_{kd},\V\Sigma_{kd})$
with $\bm\mu_{kd}=(\mu_{1kd},\ldots,\mu_{L_2kd})^\top$
and $\V\Sigma_{kd}=\mathrm{diag}(\sigma^2_{1kd},\ldots,\sigma^2_{L_2kd})$.

The free variational parameters are collected as
\[
\Theta_V
=
\left(
\{\boldsymbol\mu_r^{(j)},\boldsymbol\Sigma_r^{(j)}\}_{j=1}^{J},
\{\mu_{\ell kd},\sigma_{\ell kd}\}_{\ell,k,d}
\right).
\]
The responsibilities $\rho_i^{(j)}$ are excluded because they are updated in closed
form below rather than optimized as free parameters.

{Under the variational family above, a variational lower bound on the
target-wise composite criterion in~\eqref{model:full}, hereafter referred to as the
composite ELBO, can be written as}
\begin{equation}
\label{eq:ELBO_full}
\begin{aligned}
\mathcal{L}
&=\sum_{j=1}^J \Bigl\{
\mathbb{E}_{p(\V f^{(j)}\mid \V r^{(j)},\V W_j,\Theta_j)q(\V r^{(j)})\,q(\V W_j)\,q(\V v^{(j)})}
\Bigl[\log p\!\bigl(\V y^{(j)} \mid \V v^{(j)},\V f^{(j)},\Theta_j\bigr)\Bigr]\\
&\quad\quad+\;
\mathbb{E}_{q(\V W_j)\,q(\V v^{(j)})}
\Bigl[\log p\!\bigl(\V v^{(j)} \mid \Theta_j, \V W_j\bigr)-\log q(\V v^{(j)})\Bigr]
-\mathrm{KL}\!\bigl(q(\V r^{(j)})\;\|\;p(\V r^{(j)})\bigr)\Bigr\}\\
&\quad-\;\mathrm{KL}\!\bigl(q(\V R)\;\|\;p(\V R\mid\Theta_W)\bigr),
\end{aligned}
\end{equation}
where $q(\V W_j)=\E_{q(\V R)}[\,p(\V W_j\mid \V R)\,]$.

To enable efficient gradient-based optimization, we derive a computable closed-form for~\eqref{eq:ELBO_full}. We first define the expected kernel statistics
\begin{equation*}
    \begin{aligned}
        \Psi_{1}^{(j)} := \E_{q(\V W_j)}[\V K_{fr}^{(j)}],\qquad
        \Psi_{2}^{(i,j)} := \E_{q(\V W_j)}[\V K_{rf}^{(i,j)}\V K_{fr}^{(i,j)}]
    \end{aligned}
\end{equation*}
where we denote $\V K_{fr}^{(i,j)}\in \mathbb{R}^{1\times L_1}$ for the $i$-th row of $\V K_{fr}^{(j)}${,} and accordingly $\V K_{rf}^{(i,j)} \triangleq (\V K_{fr}^{(i,j)})^\top \in \mathbb{R}^{L_1\times 1}$. For each local region $j$ and training neighbor $i$, we define the following auxiliary scalars to represent the uncertainty propagation through the latent layers:
\[
Q_{j,i} := \frac{(y_i^{(j)}-\mu_{m(j)})^2-2(y_i^{(j)}-\mu_{m(j)})\zeta_{j,i}+A_{j,i}+B_{j,i}}{2{\sigma_{\varepsilon,j}^{2}}},
\]
where $\zeta_{j,i}$ denotes the $i$th element of $\Psi_{1}^{(j)}(\V K_{r}^{(j)})^{-1}\V\mu_r^{(j)}$, $A_{j,i} := a_{m(j)}-\tr\!\Big((\V K_{r}^{(j)})^{-1}\Psi_{2}^{(i,j)}\Big)$, and $B_{j,i} := \tr\!\Big((\V K_{r}^{(j)})^{-1}\Psi_{2}^{(i,j)}(\V K_{r}^{(j)})^{-1} \big(\V\mu_r^{(j)}\V\mu_r^{(j)\top}+\V\Sigma_r^{(j)}\big)\Big)$.

Combining these with the expected log-likelihood of the local gating function, we define the local log-evidence components $S_1^{i,j}$ and $S_2^{i,j}$:
\[
\begin{aligned}
S_1^{i,j} &:= -\tfrac12\log(2\pi{\sigma_{\varepsilon,j}^{2}})-Q_{j,i} +\E_{q(\V W_j)}\log\pi\!\big(\V\nu_j^\top[1,\V W_j\V x_i^{(j)}]\big), \\
S_2^{i,j} &:= -\log u_j +\E_{q(\V W_j)}\log\!\big(1-\pi(\V\nu_j^\top[1,\V W_j\V x_i^{(j)}])\big).
\end{aligned}
\]
For fixed values of the remaining variational and model parameters, the
coordinate-optimal responsibility is
\[
\rho_i^{(j)}
=
\frac{\exp(S_1^{i,j})}
{\exp(S_1^{i,j})+\exp(S_2^{i,j})}.
\]
Substituting this optimal responsibility into~\eqref{eq:ELBO_full} yields the
following computable form of the composite ELBO:
\begin{equation}\label{eq:closed form of elbo}
\begin{aligned}
\mathcal{L}
= &\sum_{j=1}^J\sum_{i\in\mathcal{D}_n^{(j)}}
\log\!\Big(\exp(S_1^{i,j})+\exp(S_2^{i,j})\Big)
{\;-\;\sum_{j=1}^J \mathrm{KL}\!\big(q(\V r^{(j)})\,\|\,p(\V r^{(j)})\big)}
\;\\&{-}\;
\sum_{k=1}^{Q}\sum_{d=1}^{D}\frac12\Big[
\log\frac{|\V K_{R}^{(k)}|}{|\V\Sigma_{kd}|}
-L_2+\tr\!\big((\V K_{R}^{(k)})^{-1}\V\Sigma_{kd}\big)
+\bm\mu_{kd}^\top(\V K_{R}^{(k)})^{-1}\bm\mu_{kd}
\Big],
\end{aligned}
\end{equation}
where $K_R^{(k)}$ and $(\mu_{kd}, \Sigma_{kd})$ are defined in equations~(\ref{p(R)}) and~(\ref{eq:def of q}), respectively. {Detailed derivations} are provided in Appendix~A.

In practice, we maximize $\mathcal{L}$ in~\eqref{eq:closed form of elbo}
with respect to the variational parameters $\Theta_V$ and the model parameters
$\Theta$.
Following \citep{aueb2013variational},
the global kernel hyperparameters are optimized
via a Type-II maximum likelihood (empirical Bayes) approach,
which has been shown to yield robust and computationally efficient performance
for Mahalanobis-type Gaussian process models.

\begingroup

\subsection{Prediction}

After variational training, prediction is obtained by marginalizing the local latent
variables under their variational posterior. For the $j$th target location
$\V x_*^{(j)}$, the variational posterior predictive distribution of the latent
in-regime response is defined as
\begin{equation}\label{eq:full_variational_prediction}
\begin{aligned}
    q\!\left(f_*^{(j)}\mid\mathcal D\right)
    =
    \sum_{\V v^{(j)}\in\{0,1\}^{n}}
    \int\!\!\int
    &p\!\left(
        f_*^{(j)}
        \mid
        \V r^{(j)},
        \V W_j,
        \V v^{(j)},
        \Theta_j
    \right)
    \\
    &\times
    q\!\left(\V v^{(j)}\right)
    q\!\left(\V r^{(j)}\right)
    q\!\left(\V W_j\right)
    \,d\V r^{(j)}\,d\V W_j .
\end{aligned}
\end{equation}

Conditional on the local inducing variables $\V r^{(j)}$ and the projection
matrix $\V W_j$, the target latent response is independent of the neighboring
membership indicators:
\[
    f_*^{(j)}
    \perp
    \V v^{(j)}
    \mid
    \V r^{(j)},\V W_j,\Theta_j.
\]
Consequently,
\[
    p\!\left(
        f_*^{(j)}
        \mid
        \V r^{(j)},\V W_j,\V v^{(j)},\Theta_j
    \right)
    =
    p\!\left(
        f_*^{(j)}
        \mid
        \V r^{(j)},\V W_j,\Theta_j
    \right).
\]
Using the variational factorization in \eqref{eq:def of q} and
$\sum_{\V v^{(j)}}q(\V v^{(j)})=1$, equation
\eqref{eq:full_variational_prediction} therefore reduces to
\begin{equation}\label{eq:marginalized_variational_prediction}
    q\!\left(f_*^{(j)}\mid\mathcal D\right)
    =
    \int\!\!\int
    p\!\left(
        f_*^{(j)}
        \mid
        \V r^{(j)},\V W_j,\Theta_j
    \right)
    q\!\left(\V r^{(j)}\right)
    q\!\left(\V W_j\right)
    \,d\V r^{(j)}\,d\V W_j .
\end{equation}

Let $\V K_r^{(j)}$ denote the covariance matrix of the fixed local inducing
inputs
$\{\tilde{\V z}_{\ell}^{(j)}\}_{\ell=1}^{L_1}$, as defined in
Section~\ref{sec:inference}. For a given projection matrix $\V W_j$, define the
covariance vector between the projected target input and the local inducing
inputs by
\[
    \V K_{*r}^{(j)}(\V W_j)
    =
    \left[
        \sqrt{a_{m(j)}}
        C\!\left(
            \left\|
                \V W_j\V x_*^{(j)}
                -
                \tilde{\V z}_{\ell}^{(j)}
            \right\|_2
        \right)
    \right]_{\ell=1}^{L_1}
    \in\mathbb R^{1\times L_1},
\]
and let
\[
    \V K_{r*}^{(j)}(\V W_j)
    =
    \V K_{*r}^{(j)}(\V W_j)^\top,
    \qquad
    K_{**}^{(j)}
    =
    a_{m(j)}C(0).
\]

Conditioned on $\V r^{(j)}$ and $\V W_j$, the sparse GP predictive
distribution for the latent in-regime response at $\V x_*^{(j)}$ is
\begin{equation}\label{eq:conditional_sparse_prediction}
\begin{aligned}
    p\!\left(
        f_*^{(j)}
        \mid
        \V r^{(j)},\V W_j,\Theta_j
    \right)
    =
    \mathcal N\!\Bigg(
        &\mu_{m(j)}+\V K_{*r}^{(j)}(\V W_j)
        \big(\V K_r^{(j)}\big)^{-1}
        \V r^{(j)},
        \\
        &K_{**}^{(j)}
        -
        \V K_{*r}^{(j)}(\V W_j)
        \big(\V K_r^{(j)}\big)^{-1}
        \V K_{r*}^{(j)}(\V W_j)
    \Bigg).
\end{aligned}
\end{equation}

Although \eqref{eq:conditional_sparse_prediction} is Gaussian for fixed
$\V r^{(j)}$ and $\V W_j$, the marginal predictive distribution in
\eqref{eq:marginalized_variational_prediction} is generally non-Gaussian because
the kernel vector depends nonlinearly on the random projection matrix $\V W_j$.
We therefore approximate this distribution by a Gaussian whose mean and variance
match its first two moments.

To evaluate these moments, define the target--inducing expected kernel statistics
\begin{equation}\label{eq:pred_expected_kernel_stats}
    \begin{aligned}
    \Psi_{1,*}^{(j)}
    &:=
    \mathbb E_{q(\V W_j)}
    \!\left[
        \V K_{*r}^{(j)}(\V W_j)
    \right],
    \\
    \Psi_{2,*}^{(j)}
    &:=
    \mathbb E_{q(\V W_j)}
    \!\left[
        \V K_{r*}^{(j)}(\V W_j)
        \V K_{*r}^{(j)}(\V W_j)
    \right].
    \end{aligned}
\end{equation}
By the law of total expectation, the predictive mean is
\begin{equation}\label{eq:pred_mean}
    \widehat\mu_*^{(j)}
    :=
    \mathbb E_q\!\left[f_*^{(j)}\right]
    =
    \mu_{m(j)}+\Psi_{1,*}^{(j)}
    \big(\V K_r^{(j)}\big)^{-1}
    \bm\mu_r^{(j)} .
\end{equation}
By the law of total variance, the corresponding latent predictive variance is
\begin{equation}\label{eq:pred_latent_var}
\begin{aligned}
    \widehat s_{f,*}^{2(j)}
    :=
    \operatorname{Var}_q\!\left(f_*^{(j)}\right)
    =
    K_{**}^{(j)}
    &-
    \operatorname{tr}\!\left[
        \big(\V K_r^{(j)}\big)^{-1}
        \Psi_{2,*}^{(j)}
    \right]
    \\
    &+
    \operatorname{tr}\!\left[
        \big(\V K_r^{(j)}\big)^{-1}
        \Psi_{2,*}^{(j)}
        \big(\V K_r^{(j)}\big)^{-1}
        \left(
            \V\Sigma_r^{(j)}
            +
            \bm\mu_r^{(j)}
            \bm\mu_r^{(j)\top}
        \right)
    \right]
    \\
    &-
    \left(\widehat\mu_*^{(j)}-\mu_{m(j)}\right)^2 .
\end{aligned}
\end{equation}
The moment-matched approximation to the latent predictive distribution is
therefore
\begin{equation}\label{eq:pred_latent_dist}
    q\!\left(f_*^{(j)}\mid\mathcal D\right)
    \approx
    \mathcal N\!\left(
        \widehat\mu_*^{(j)},
        \widehat s_{f,*}^{2(j)}
    \right).
\end{equation}

For prediction of the noisy response, the local observation-noise variance is
added to the latent predictive variance:
\begin{equation}\label{eq:pred_noisy_var}
    \widehat s_{y,*}^{2(j)}
    =
    \widehat s_{f,*}^{2(j)}
    +
    {\widehat\sigma_{\varepsilon,j}^{2}} .
\end{equation}
The resulting predictive distribution is
\begin{equation}\label{eq:pred_dist_single}
    y_*^{(j)}\mid\mathcal D
    \approx
    \mathcal N\!\left(
        \widehat\mu_*^{(j)},
        \widehat s_{y,*}^{2(j)}
    \right).
\end{equation}

The
corresponding derivations are provided in Appendix~A.
\endgroup

\begin{remark}[Effect of neighborhood overlap on uncertainty]\label{rem:overlap_uq}
Each factor of the composite likelihood in Remark~\ref{rem:composite} uses only its own
neighborhood, and the predictive
distribution~\eqref{eq:marginalized_variational_prediction} conditions on $\V r^{(j)}$ and
$\V W_j$ alone. Overlap therefore leaves inference within a single target unaffected; what it
affects is the shared projection layer, since an observation lying in several neighborhoods
contributes more than once to $q(\V R)$ and hence to $q(\V W_j)$. As is typical for composite
likelihoods, such repeated contributions can make the projection-layer posterior too
concentrated. We assess the resulting predictive uncertainty empirically through $90\%$
coverage (Cov90) in Sections~\ref{sec:simulations} and~\ref{sec:real}.
\end{remark}

\begingroup
\begin{algorithm}[htbp]
\caption{DJGP: Variational Training \& Analytic Prediction}
\label{alg:DJGP}
\begin{algorithmic}[1]
\REQUIRE Target-specific local datasets \(\{(\mathcal D_n^{(j)},\V{x}^{(j)}_*)\}_{j=1}^J\), initial values of the
variational parameters $\Theta_V$, fixed local inducing inputs
$\{ \tilde{\V z}_{\ell}^{(j)}: \ell = 1,\ldots,L_1,\; j = 1,\ldots,J \}$,
fixed global inducing inputs
$\{ \tilde{\V x}_{\ell}: \ell = 1,\ldots,L_2 \}$,
model parameters $\Theta$, learning rate \(\eta\), and max iterations \(S\)
\ENSURE Posterior approximations and predictive distribution at test points

\FOR{\(s=1\) to \(S\)}
    \STATE Compute the variational projection posteriors
    \(q(\V W_j)=\E_{q(\V R)}[\,p(\V W_j\mid\V R,\Theta_W)\,]\) from \(q(\V R)\) and the
    sparse GP conditional associated with the prior \eqref{p(R)}
    \STATE Compute the expected local kernel statistics
    \(\{\Psi_{1}^{(j)},\Psi_{2}^{(i,j)}\}_{i,j}\) as defined in
    Section~\ref{sec:inference}
    \STATE Compute $S_1^{i,j}$ and $S_2^{i,j}$ and set
    $\rho_i^{(j)}=\exp(S_1^{i,j})/[\exp(S_1^{i,j})+\exp(S_2^{i,j})]$;
    \STATE Compute \(\mathcal{L}\) in~\eqref{eq:closed form of elbo} and its gradients w.r.t.\
    all optimized variational and model parameters \((\Theta_V,\Theta)\)
    \STATE Update parameters by gradient ascent:
    \({(\Theta_V,\Theta)\leftarrow(\Theta_V,\Theta)
    +\eta\,\nabla_{(\Theta_V,\Theta)}\mathcal{L}}\)
    \STATE Enforce positivity constraints on variance and length-scale parameters
    \STATE \textbf{If} {\(\mathcal L\) has converged} \textbf{then} \textbf{break}
\ENDFOR

\STATE \textbf{Analytic prediction:}
\FOR{{each target location} \(j\)}
    \STATE Compute the projection posterior \(q(\V W_j)\) from \(q(\V R)\) and the sparse GP
    conditional
    \STATE Compute the target-inducing expected kernel statistics
    \(\{\Psi_{1,*}^{(j)},\Psi_{2,*}^{(j)}\}\) as in
    \eqref{eq:pred_expected_kernel_stats}
    \STATE Compute the analytic predictive mean and variance
    \(\{ \widehat\mu_{*}^{(j)}, \widehat s_{y,*}^{2(j)} \}\)
    as in \eqref{eq:pred_mean}--\eqref{eq:pred_noisy_var}
\ENDFOR
\RETURN \(\{ \widehat\mu_{*}^{(j)}, \widehat s_{y,*}^{2(j)} \}_{j=1}^J\)
\end{algorithmic}
\end{algorithm}
\endgroup

\section{Theoretical Results}\label{sec:theorem}

\begingroup

\spacingset{1.35}

We next provide a finite-sample analysis that clarifies why the proposed
local-projection construction can mitigate the high-dimensional difficulty of JGP.
Although DJGP operates on inputs in $\mathbb R^D$, its local prediction is based on a
learned $K$-dimensional representation. A natural question is therefore whether the
prediction error of DJGP can be related to that of an {oracle GP operating on
ideal local latent coordinates and the correct local regime}.

For a generic target $\V x_*$, let $\mathcal D_n$ denote the indices of its $n$
nearest training neighbors and define the corresponding neighborhood radius as
\[
    \rho_r(\V x_*)
    :=
    \max_{i\in\mathcal D_n}
    \|\V x_i-\V x_*\|.
\]
Because the local projection is identifiable only up to the latent scale and row
orientation, we fix one representative coordinate convention and write
$\V W_*=g'(\V x_*)$. Under this convention, the smooth latent map admits the local
expansion
\[
    g(\V x_i)-g(\V x_*)
    =
    \V W_*(\V x_i-\V x_*)
    +
    O\!\left(\|\V x_i-\V x_*\|^2\right).
\]
Here $K$ denotes the intrinsic dimension of the latent map $g$, while $Q$ is the projection dimension chosen by DJGP. For the theoretical analysis in Theorem~\ref{thm:6}, we assume that the chosen dimension $Q$ is equal to the intrinsic dimension $K$. Accordingly, Theorem~\ref{thm:6} does not characterize the effects of under- or over-specifying $Q$. In practice, $Q$ is treated as a model-selection parameter and need not equal $K$, as examined empirically in Section~D.3.3 of the Online Supplementary Material. {For the theoretical analysis, let $m_*$ denote the region
containing $g(\V x_*)$, and define the oracle same-region subset $\mathcal D_*$
and its size $n_*$}
\[
    \mathcal D_*
    :=
    \{i\in\mathcal D_n:g(\V x_i)\in\mathcal Z_{m_*}\},
    \qquad
    n_*:=|\mathcal D_*|
\]

Our target is the prediction risk
\[
    \mathcal R
    :=
    \mathbb E
    \left[
        \left\{
            \widehat\mu_*-f(g(\V x_*))
        \right\}^2
    \right],
\]
where $\widehat\mu_*$ is the actual analytic DJGP predictive mean in
\eqref{eq:pred_mean}, and the expectation is over both the test point and the
training data.

\begin{assumption}[Smooth latent map]\label{ass:g}
The latent map $g$ is twice continuously differentiable in the local neighborhood,
with uniformly bounded Hessian: there exists $M_g<\infty$ such that
\[
    \|\nabla^2 g(\V x)\| \le M_g.
\]
\end{assumption}

\begin{assumption}[Within-region regularity]\label{ass:f}
For each latent region $\mathcal Z_m$, let $\mu_m$ denote its GP mean and let
$c_m$ denote its regional covariance kernel. We assume
\[
    f_m-\mu_m\in\mathcal H_{c_m},
    \qquad
    \|f_m-\mu_m\|_{\mathcal H_{c_m}}\le B_f,
\]
and that the corresponding regional function is locally Lipschitz:
\[
    |f_m(z)-f_m(z')|
    \le
    L_f\|z-z'\|_2
\]
for nearby points belonging to the same regional component or its local extension.
The sufficient canonical-metric condition is stated in Appendix~B.3.
\end{assumption}

\begin{assumption}[Local gate accuracy]\label{ass:margin}
For each neighbor $i\in\mathcal D_n$, define the oracle in-region label
\[
    r_i^*
    :=
    \mathbb I\{g(\V x_i)\in\mathcal Z_{m_*}\},
\]
where $m_*$ denotes the region containing $g(\V x_*)$.
Let $\widehat r_i\in\{0,1\}$ denote the learned hard-gate label, obtained by
thresholding the marginalized logistic gate probability at $1/2$.
Define
\[
    I_i:=\mathbb I\{\widehat r_i\neq r_i^*\}.
\]
We assume
\[
    \mathbb E
    \left[
        \frac{1}{n}
        \sum_{i\in\mathcal D_n} I_i
    \right]
    \le \epsilon_n.
\]
Thus, $\epsilon_n$ measures the expected fraction of local neighbors whose learned
regime assignment differs from the oracle assignment.
\end{assumption}

The analysis compares the implemented DJGP predictor with an idealized full
local-GP reference that allows the effects of gating, projection, and local geometry
to be isolated. Let $\mathcal E_{\mathrm{approx}}$ denote their squared prediction
difference, as defined by the exact decomposition in Appendix~B.1.  The same
subsection states the bounded-response and centered finite-variance noise conditions;
Appendix~B.4 gives the corresponding high-probability formulation under Gaussian or
sub-Gaussian noise.  We write $n_*$ for the oracle same-region count.

\begin{theorem}[DJGP risk decomposition]\label{thm:6}
Under Assumptions~\ref{ass:g}--\ref{ass:margin}, the bounded-response and noise
conditions in Appendix~B.1, the {subset-stability condition in Appendix~B.4
(Assumption~A2)}, and the bounded-domain,
kernel-regularity, and conditional-Nystr\"om conditions in Appendix~B.5
(Assumptions~A3--A4 and A6),
\begin{align}
\mathcal R
\lesssim\;&
\underbrace{
    \mathcal E_{\mathrm{approx}}
}_{\text{approximation}}
+
\underbrace{
    KD\,T(L_2)
    +
    \mathbb E
    \|\overline{\V W}-\V W_*\|_F^2
}_{\text{projection}}
\nonumber\\
&\quad+
\underbrace{
    \mathbb E[\rho_r(\V x_*)^4]
    +
    \sigma_\varepsilon^2
}_{\text{local approximation}}
+
\underbrace{
    \epsilon_n^{2/3}
}_{\text{gating}}
+
\underbrace{
    \mathrm{GP}_{\mathrm{oracle}}
    (n_*,K)
}_{\text{oracle regression}} .
\label{eq:overall-risk-final}
\end{align}
Here $T(L_2)$ measures the finite-inducing approximation of the projection GPs,
$\overline{\V W}$ is the learned conditional-mean projection, and
$\mathrm{GP}_{\mathrm{oracle}}(n_*,K)$ is the prediction risk of an
{oracle GP operating on the ideal $K$-dimensional local coordinates and the
$n_*$ same-region neighbors}.  The constants hidden by $\lesssim$ may depend
on the fixed local size $n$, the bounded-response constant, and the stated kernel and
noise regularity constants; in particular, Appendix~B.4 gives the gating contribution
as $O(B_y^2\epsilon_n^{2/3})$.
\end{theorem}

Theorem~\ref{thm:6} shows that, when the approximation, gating, and
projection errors are controlled, DJGP approaches an oracle GP operating directly
in the $K$-dimensional latent space. Thus, the remaining nonparametric regression
difficulty is governed primarily by the intrinsic dimension $K$, rather than the
ambient dimension $D$. The ambient dimension enters through the difficulty of
recovering the local projection, represented by the projection terms in
\eqref{eq:overall-risk-final}.

The decomposition also explains why locally linear projections can be effective even
when the underlying dimension-reduction map $g$ is globally nonlinear. By
Assumption~\ref{ass:g}, the local Taylor remainder is
$O(\rho_r(\V x_*)^2)$, leading to a squared prediction-error contribution of order
$O(\rho_r(\V x_*)^4)$. Under Assumption~A5 in Appendix~B.5,
$\mathbb E[\rho_r(\V x_*)^4]\to0$ as $N\to\infty$, so this geometric
approximation error vanishes.

\begin{remark}[Oracle regression rate]
Under the additional regional-design conditions in Appendix~B.6, {Corollary~2},
\[
    \mathrm{GP}_{\mathrm{oracle}}(n_*,K)
    \lesssim
    B_f^2
    \frac{\gamma_{n_*+1}}
         {n_*+1},
\]
where $\gamma_t$ is the maximum information gain of the latent kernel.
Consequently, the oracle term is
\[
    O\!\left(
        B_f^2
        \frac{(\log n_*)^{K+1}}
             {n_*}
    \right)
\]
for the squared-exponential kernel and
\[
    \widetilde O\!\left(
        B_f^2
        n_*^{-2\nu_M/(2\nu_M+K)}
    \right)
\]
for a Mat\'ern$(\nu_M)$ kernel
\citep{vakili2021information}.
\end{remark}

The result is intended as a finite-sample risk decomposition relative to an
oracle local-GP reference rather than a general posterior-contraction result.
Appendix~B.1 defines the reference
predictors and exact decomposition, Appendix~B.2 states the aligned supplemental
theorem, Appendices~B.3--B.6 develop the local-linearization, gating, projection, and
oracle-regression bounds, respectively, and Appendix~B.7 combines them.

\endgroup

\section{Synthetic Dataset Experiments}\label{sec:simulations}
We assess the performance of DJGP on synthetic datasets, comparing against several baseline methods\footnote{The Python implementation and a lightweight end-to-end
reproducibility benchmark are available at
\url{https://github.com/crushonyfg/DJGP_pub}. See Section~D.7 of the supplement
for the fixed reproduction protocol and reference results.}. 
We report root mean squared error (RMSE), continuous ranked probability score (CRPS), {and empirical 90\% predictive interval coverage (Cov90)} as evaluation metrics; we favor CRPS over negative log predictive density (NLPD) because of NLPD’s sensitivity to outliers. For a Gaussian predictive distribution $\mathcal{N}(\mu_j,\sigma_j^2)$ at the $j$th test site ($j=1,{\ldots},J$) and the test response $y_*^{(j)}${, where $\sigma_j^2$ denotes the predictive variance $\widehat s_{y,*}^{2(j)}$ of~\eqref{eq:pred_noisy_var}}, the metrics are defined as
	$\mathrm{RMSE}
	=  \sqrt{\frac{1}{J}\sum_{j=1}^J (y_*^{(j)} - \mu_j)^2}\,$ and $\mathrm{CRPS}
	= 1/J \sum_{j=1}^J \sigma_j \Bigl[\,\tilde{y}_j \bigl(2\Phi(\tilde{y}_j)-1\bigr)
+2\phi(\tilde{y}_j)-1/\sqrt\pi\Bigr],$ with
$\tilde{y}_j = (y_*^{(j)} - \mu_j)/\sigma_j$,
where \(\Phi(\cdot)\) and \(\phi(\cdot)\) are the standard normal CDF and PDF, respectively. {The empirical 90\% coverage is defined as
$
    \mathrm{Cov90}
    =
    \frac{1}{J}
    \sum_{j=1}^{J}
    \mathbb I
    \left\{
        y_*^{(j)}
        \in
        \left[
            \mu_j-z_{0.95}\sigma_j,\,
            \mu_j+z_{0.95}\sigma_j
        \right]
    \right\},
$
where $z_{0.95}=\Phi^{-1}(0.95)\approx 1.645$. A well-calibrated predictive distribution
should have $\mathrm{Cov90}$ close to $0.90$.}

\paragraph{Benchmark Methods}

The first baseline is the Jump Gaussian Process (\textbf{JGP}) proposed by \citet{park2022jump},
implemented with classification EM and linear partition boundaries.
We restrict JGP to linear separators to avoid overfitting in high-dimensional settings and to ensure a fair comparison
with DJGP.
The second baseline is \textbf{JGP-PCA}, which applies principal component analysis to
reduce the input dimension before fitting a standard JGP on the projected features.
The third baseline is \textbf{JGP-SIR}, which applies sliced inverse regression \citep{li1991sliced}
to reduce the input dimension before fitting a standard JGP on the projected features.
Fourth, we include a two-layer, doubly-stochastic Deep Gaussian Process (\textbf{DGP}) {\citep{salimbeni2017doubly}, implemented in GPyTorch~\citep{gardner2018gpytorch}}.
This model can also be interpreted as a Gaussian Process Latent Variable Model (GP-LVM).
Since both DGP and DMGP are global, smooth, inductive deep-GP models, for brevity we report only DGP---the more
commonly used global deep-GP baseline---in the main results. {As shown in Section~D.6 of the supplement, DMGP and DGP show broadly comparable predictive accuracy.} \begingroup
Lastly, we include \textbf{NGBoost} as a strong non-GP probabilistic baseline
\citep{duan2020ngboost}. Standard tree-based or neural-network regressors are primarily
designed for point prediction, and their predictive variances are often obtained indirectly
through bootstrap, dropout, or ensemble heuristics. Such uncertainty estimates can be
unstable because they are not tied to a proper probabilistic scoring objective. NGBoost
instead learns a parametric predictive distribution by natural-gradient boosting under a
proper scoring rule. In our experiments, we use the official \texttt{ngboost} package and fit
a Gaussian predictive distribution with the logarithmic score,
$
    y\mid \V x \sim \mathcal N\!\left(\mu(\V x),\sigma^2(\V x)\right),
$
where both the location and scale parameters are learned as functions of the input. This
makes NGBoost a competitive non-GP baseline for evaluating probabilistic prediction and
uncertainty quantification. Finally, we include Bayesian Additive Regression Trees (\textbf{BART})
\citep{chipman2010bart} as a flexible Bayesian tree-ensemble baseline. BART models the
regression function as a sum of weak regression trees and uses a regularizing prior to
shrink the contribution of each tree. Unlike NGBoost, which directly optimizes a
parametric predictive distribution, BART obtains uncertainty through its posterior
predictive distribution induced by Bayesian backfitting and posterior sampling. We include
BART to compare against a widely used Bayesian nonparametric regression method that can
capture nonlinear interactions while providing posterior predictive uncertainty.

\endgroup

\paragraph{Synthetic Dataset{} Construction}
Synthetic datasets are generated using a two-stage framework. In the first stage, we construct a latent feature space \(\mathcal{Z}\) of dimension \(K\) and define the response relationship as \(y = f(\boldsymbol{z}) + \epsilon\). In the second stage, we apply dimensionality expansion by simulating the inverse of a smooth projection \(g^{-1} : \mathcal{Z} \rightarrow \mathcal{X}\), lifting the latent dimension \(K\) to an observed dimension \(D\). The resulting datasets therefore have \(D\)-dimensional inputs.

\begin{figure}[ht!]
    \centering
    \begin{subfigure}[t]{0.46\linewidth}
        \centering
        \includegraphics[height=3.6cm]{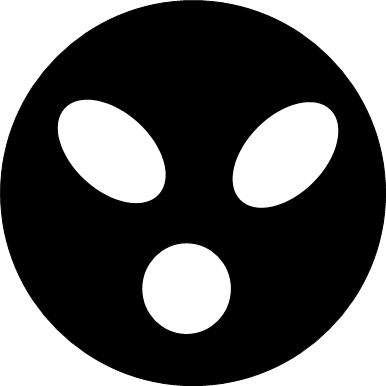}
        \caption{{L2 dataset.}}
    \end{subfigure}
    \hfill
    \begin{subfigure}[t]{0.46\linewidth}
        \centering
        \includegraphics[height=3.6cm]{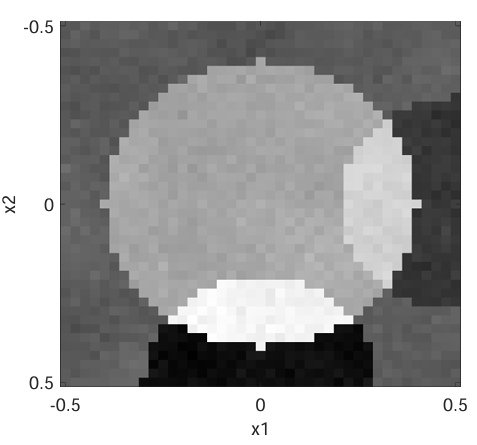}
        \caption{{LH dataset.}}
    \end{subfigure}
    \caption{Illustrations of response surfaces over 2D latent spaces. (a) A noiseless ground-truth regression surface with sharp transitions {from the L2 dataset}. (b) {An example realization from the high-dimensional LH dataset, projected onto the 2D latent domain $[-0.5,0.5]^2$}.}
    \label{fig:latent_combined}
\end{figure}

The first stage generates $\{(\boldsymbol{z}, y), y = f(\V{z}) + \epsilon\}$, using the following two scenarios:
\begin{sloppypar}%
\begin{itemize}
    \item \textbf{L2 Dataset $K=2$:} A two-dimensional synthetic dataset defined on \([-0.5,0.5]^2\), partitioned into multiple regions {by a fixed pixel-wise region-label template, as illustrated in
Figure~\ref{fig:latent_combined}(a) and provided with the released code.
The image is binarized into two pixel labels, and each latent input is
assigned the label of its corresponding pixel.} Within each region \(m\), the response surface is drawn from an independent GP with mean \(\mu_m \in \{-5,5\}\) and squared-exponential covariance
    $
    c(\boldsymbol z,\boldsymbol z';\V\theta_m)
    = \theta_{m1}\exp\Bigl\{-\tfrac{1}{\theta_{m2}}(\boldsymbol z-\boldsymbol z')^\top(\boldsymbol z-\boldsymbol z')\Bigr\},
    $
    where \(\theta_{m1}=9\) and \(\theta_{m2}=0.05\) with Gaussian noise \(({\sigma_{\varepsilon}^{2}}=1)\). A total of 1,200 points are sampled uniformly, with 1,000 used for training and 200 for testing.

    \item \textbf{LH Dataset $K=4,5,7$:} A higher-dimensional dataset defined on \([-0.5,0.5]^K\) for \(K = 4,5,7\). We define \(K+1\) partitioning functions:
    \[
    f_0(\boldsymbol z)=\sum_{i=1}^K z_i^2 - 0.4^2,
    \quad
    f_j(\boldsymbol z)=\sum_{i=1}^K z_i^2 - z_j^2 + (z_j + r_j \cdot 0.5)^2 - 0.3^2,\quad j=1,\dots,K,
    \]
    where \(r_j \sim \mathrm{Uniform}\{\pm1\}\). Each \(f_j\) partitions the domain into two regions, and the region index is defined as
    {
    \[
     \sum_{j=0}^K 2^j\,\mathbb I_{\mathcal Z_{j,+}}(\boldsymbol z),
     \qquad\text{where}\quad
     \mathcal Z_{j,+}
     =
     \left\{
     \boldsymbol z\in[-0.5,0.5]^K:
     f_j(\boldsymbol z)\geq 0
     \right\}.
    \]
    }%
    Responses are sampled from region-dependent GPs with mean \(\mu_m \sim {\mathrm{Uniform}\bigl(\{-13.5m,\,13.5m\}\bigr)}\) and the same covariance as above. Gaussian noise \(\epsilon \sim \mathcal N(0,4)\) is added. We generate \(N\) training samples uniformly. {200 test inputs are generated
by rejection sampling: candidate points are first sampled uniformly from
$[-0.5,0.5]^K$, and a candidate is retained only if its boundary score
{
\[
    d(\V z)
    =
    \min_{0\le j\le K}
    \frac{|f_j(\V z)|}{\|\nabla f_j(\V z)\|_2}
\]
}%
is no larger than the threshold $\delta=0.05${,} following the setting in JGP~\citep{park2022jump}.}
\end{itemize}
\end{sloppypar}%

To emulate high-dimensional inputs, the second stage lifts $\V{z} \in \mathbb{R}^K$ to high-dimensional inputs $\V{x} \in \mathbb{R}^D$ using random projection (RP), random Fourier features (RF), polynomial expansion (PE), autoencoder-based expansion (AE), {or manifold embedding expansion (ME)}. Details are provided in Appendix~C.
Each dataset is defined by a combination of first- and second-stage choices. {Table~\ref{tab:all_metrics_compact} summarizes the resulting configurations, including the expansion type, latent dimension $K$, and observed dimension $D$; unless otherwise stated, each configuration uses $N=1000$ training points and $J=200$ held-out test points.} {This yields 20 synthetic datasets in total.}

\paragraph{Implementation Details}
For each dataset setting, we repeat the experiment 20 times to account for randomness and report averaged results. Hyperparameters are tuned via five-fold cross-validation on the training set for the first run of each setting; the selected values are then reused in subsequent runs to reduce computational cost. {Specifically, in the experiments, candidate initializations for $\V R_\ell$ are constructed from PCA,
SIR, partial least squares regression (PLS), and five random orthonormal projections. We
train DJGP from each candidate initialization on the validation split and select the
initialization with the best validation CRPS. The latent dimension \(Q\) (the dimension actually used by DJGP, a modeling choice that may differ from the intrinsic, data-generating dimension $K$) is selected from \(\{3,5,7\}\), the neighborhood size is selected from \(\{25, 35\}\), and the number
of optimization iterations \(S\) is selected from \(\{200,400,600\}\). After selection, each model is retrained on the full training set and evaluated on a held-out test set.}
{The additional scaling, robustness, and tuning experiments are summarized in Section~\ref{sec:additional_analysis}, with full settings and results in Sections~D.1--D.3 of the supplement.}

\begingroup
We use $L_1=10$ local inducing points per target and set $L_2=J$, so the global
inducing inputs of the projection GP are the full set of test locations
$\tilde{\V x}_\ell=\V x_*^{(\ell)}$, $\ell=1,\ldots,J$. Given a candidate projection
$\V W_0$ (PCA, SIR, PLS, or a random orthonormal projection, as above), the global
inducing outputs $\V R$ and the projection posterior are seeded from $\V W_0$ so that the
initial local projections match $\V W_0$, while the zero-mean projection GP prior
regularizes the projections to vary smoothly across target locations. The local-anchor
reprojection and the full initialization rules are detailed in
Appendix~A.
For all methods, both the input variables and response values are
standardized using the training-set mean and standard deviation before model fitting.
Predictions are then transformed back to the original response scale before computing
RMSE, CRPS, and Cov90.
\endgroup
The fixed-versus-learned inducing-input ablation is reported in
Section~D.5 of the supplement.

\subsection{Comparison to the Baseline Methods}

{Table~\ref{tab:all_metrics_compact} provides the numerical results for each experimental configuration and summarizes the average per-configuration rank scores over all 20 configurations for RMSE, CRPS, and coverage.} For each configuration and metric, the seven methods are ranked from $1$ (best) to {$7$} (worst)---by smaller RMSE or CRPS and by Cov90 closest to $0.90$---and the reported rank score is the average of these ranks across the $20$ configurations.

\begin{table*}[ht]
\centering
\tiny
\setlength{\tabcolsep}{1.8pt}
\renewcommand{\arraystretch}{0.9}
\caption{Per-configuration results on RMSE, CRPS, and 90\% coverage.}
\label{tab:all_metrics_compact}
\resizebox{\textwidth}{!}{%
\begin{tabular}{lccc|c|c|c|c|c|c|c}
\toprule
\multicolumn{4}{c|}{\textbf{Dataset}}
& \textbf{DJGP(Ours)}
& \textbf{JGP}
& \textbf{JGP-PCA}
& \textbf{JGP-SIR}
& \textbf{DGP}
& \textbf{NGBoost}
& \textbf{BART} \\
\textbf{} & \textbf{D} & \textbf{K} & \textbf{Feat.}
& \textbf{R/C/C90}
& \textbf{R/C/C90}
& \textbf{R/C/C90}
& \textbf{R/C/C90}
& \textbf{R/C/C90}
& \textbf{R/C/C90}
& \textbf{R/C/C90} \\
\midrule

\textbf{L2} & 20 & 2 & \textbf{AE}
& 1.35/\textbf{0.50}/0.87
& 1.43/0.53/0.78
& 1.42/0.52/0.77
& 1.83/0.78/0.79
& 1.40/0.68/0.95
& 1.37/0.58/0.79
& \textbf{1.32}/0.58/\textbf{0.90} \\

& 20 & 2 & \textbf{ME}
& \textbf{1.30}/\textbf{0.50}/0.86
& 1.47/0.53/0.78
& 1.52/0.53/0.76
& 2.07/0.83/0.81
& 1.38/0.67/0.95
& 1.31/0.57/0.77
& 1.38/0.60/\textbf{0.90} \\

& 20 & 2 & \textbf{PE}
& \textbf{1.42}/\textbf{0.52}/0.86
& 1.55/0.58/0.75
& 2.30/0.97/0.76
& 2.09/0.96/0.80
& \textbf{1.42}/0.69/0.94
& 1.50/0.61/0.78
& 1.46/0.64/\textbf{0.91} \\

& 20 & 2 & \textbf{RP}
& \textbf{1.19}/\textbf{0.48}/0.86
& 1.48/0.53/0.80
& 1.53/0.53/0.77
& 1.52/0.53/0.80
& 1.81/0.93/0.96
& 1.45/0.61/0.78
& 1.38/0.60/\textbf{0.90} \\

& 20 & 2 & \textbf{RF}
& 1.58/\textbf{0.51}/0.84
& 1.46/0.54/0.77
& 1.92/0.74/0.78
& 2.12/0.90/0.80
& 1.67/0.85/0.94
& \textbf{1.42}/0.60/0.77
& 1.43/0.61/\textbf{0.91} \\

\midrule

\textbf{LH} & 20 & 4 & \textbf{AE}
& \textbf{217}/111/\textbf{0.88}
& 295/129/0.35
& 339/172/0.39
& 292/131/0.53
& 226/\textbf{110}/0.71
& 243/136/0.65
& 262/150/0.76 \\

& 20 & 4 & \textbf{ME}
& \textbf{232}/118/\textbf{0.88}
& 343/174/0.34
& 351/182/0.39
& 355/177/0.51
& 236/\textbf{117}/0.72
& 258/142/0.75
& 271/154/0.76 \\

& 20 & 4 & \textbf{PE}
& \textbf{217}/\textbf{112}/\textbf{0.88}
& 344/165/0.38
& 330/165/0.60
& 312/148/0.55
& 248/124/0.66
& 247/133/0.70
& 258/148/0.76 \\

& 20 & 4 & \textbf{RP}
& 252/\textbf{129}/\textbf{0.80}
& 340/172/0.36
& 346/174/0.41
& 353/177/0.42
& \textbf{249}/130/0.70
& 263/143/0.76
& 272/154/0.76 \\

& 20 & 4 & \textbf{RF}
& \textbf{219}/\textbf{110}/\textbf{0.87}
& 322/146/0.31
& 370/196/0.47
& 314/146/0.54
& 322/183/0.68
& 260/140/0.72
& 270/153/0.77 \\

\midrule

\textbf{LH} & 30 & 5 & \textbf{AE}
& \textbf{519}/\textbf{246}/\textbf{0.81}
& 647/296/0.24
& 672/321/0.33
& 610/264/0.43
& 525/262/0.73
& 539/272/0.60
& 577/290/0.80 \\

& 30 & 5 & \textbf{ME}
& \textbf{528}/\textbf{242}/\textbf{0.86}
& 662/311/0.31
& 712/323/0.35
& 679/323/0.38
& 659/324/0.79
& 584/280/0.71
& 591/293/0.80 \\

& 30 & 5 & \textbf{PE}
& \textbf{525}/\textbf{238}/\textbf{0.89}
& 834/478/0.28
& 629/280/0.47
& 624/274/0.47
& 650/331/0.73
& 536/263/0.67
& 561/286/0.80 \\

& 30 & 5 & \textbf{RP}
& \textbf{524}/\textbf{244}/\textbf{0.86}
& 675/329/0.31
& 676/325/0.34
& 678/321/0.33
& 661/325/0.79
& 588/282/0.71
& 593/293/0.80 \\

& 30 & 5 & \textbf{RF}
& \textbf{526}/\textbf{246}/\textbf{0.87}
& 725/368/0.24
& 686/336/0.38
& 617/269/0.40
& 623/318/0.72
& 598/294/0.66
& 591/293/0.80 \\
\midrule

\textbf{LH} & 50 & 7 & \textbf{AE}
& 1811/628/\textbf{0.89}
& 2142/812/0.25
& 1868/634/0.26
& 1843/617/0.29
& \textbf{1669}/\textbf{601}/0.79
& 2385/1005/0.52
& 2376/1007/0.83 \\

& 50 & 7 & \textbf{ME}
& 1796/615/\textbf{0.88}
& \textbf{1791}/\textbf{605}/0.23
& 1850/630/0.26
& 1828/613/0.25
& 1816/629/\textbf{0.88}
& 2388/1004/0.55
& 2388/1011/0.83 \\

& 50 & 7 & \textbf{PE}
& \textbf{1748}/629/0.87
& 2534/1117/0.43
& 1856/648/0.35
& 1813/\textbf{611}/0.36
& 1805/634/\textbf{0.86}
& 2365/1009/0.51
& 2339/1005/0.84 \\

& 50 & 7 & \textbf{RP}
& 1852/640/\textbf{0.88}
& 1839/629/0.24
& \textbf{1812}/\textbf{604}/0.29
& 1816/608/0.24
& 1820/630/\textbf{0.88}
& 2404/1009/0.54
& 2387/1011/0.83 \\

& 50 & 7 & \textbf{RF}
& \textbf{1768}/\textbf{613}/\textbf{0.88}
& 2264/883/0.28
& 1841/633/0.29
& 1841/620/0.27
& 1857/667/0.83
& 2413/1028/0.49
& 2371/1008/0.84 \\

\midrule
\multicolumn{4}{l|}{\textbf{Per-config rank score}}
& \textbf{1.58}/\textbf{1.55}/\textbf{1.38}
& 5.18/4.38/6.53
& 5.48/4.90/5.88
& 5.03/4.08/5.20
& 3.43/4.18/2.93
& 3.58/4.08/4.18
& 3.75/4.85/1.93 \\

\bottomrule
\end{tabular}%
}
\vspace{1mm}
\begin{minipage}{0.96\textwidth}
\footnotesize
\textit{Note.} Each entry is reported as R/C/C90, where R, C, and C90 denote RMSE, CRPS, and Cov90,
respectively. Lower RMSE and CRPS are better, while Cov90 is better when closer to
\(0.90\). Bold entries indicate the best or rounded-tied best method within each configuration and metric.
Per-configuration ranks are averaged across the 20 settings, with average
ranks assigned to ties; lower scores are better.
\end{minipage}
\end{table*}

\begingroup
Overall, DJGP achieves the strongest aggregate performance across the synthetic
benchmarks. As shown in Table~\ref{tab:all_metrics_compact}, DJGP obtains the best average
rank scores for RMSE, CRPS, and Cov90 across the 20 configurations{, giving the
strongest overall accuracy--calibration trade-off}.

Compared with JGP and its two-stage dimension-reduced variants, DJGP benefits from
learning the projection jointly with the local GP/JGP objective. In the higher-dimensional
settings, especially when \(D=30\) or \(D=50\), PCA- and SIR-based preprocessing often
improves over vanilla JGP, suggesting that dimension reduction is indeed useful for
stabilizing local modeling in high-dimensional input spaces. DJGP generally
outperforms both JGP-PCA and JGP-SIR, indicating that optimizing the projection for the
downstream local probabilistic model is more effective than learning it as a separate
preprocessing step.
In the lower-dimensional ($D=20$) settings, the benefit of external dimension reduction is less consistent. JGP-PCA and JGP-SIR do not always improve over JGP{; plug-in dimension reduction may discard useful information when the original JGP can still handle comparatively modest input dimensions}. In contrast, DJGP remains more robust in these cases by adapting the local projection while preserving the local JGP structure. In two of the highest-dimensional ($D=50$) configurations, JGP attains a slightly smaller CRPS than DJGP, but it does so with severe undercoverage (Cov90 near $0.2$ against $0.85$ or above for DJGP), so the lower CRPS should not be read as evidence of better uncertainty quantification.

The comparison with NGBoost and DGP is also informative, since both serve as strong global probabilistic baselines. NGBoost is competitive in RMSE and provides a strong non-GP probabilistic regression baseline. However, its predictive intervals often under-cover{, so its competitive RMSE coexists with substantial undercoverage and should not be read as evidence of well-calibrated uncertainty. NGBoost learns a global conditional distribution without an explicit kernel-based notion of local similarity or smoothness, which is one possible explanation for this behavior in locally sparse regions}. DGP provides more competitive uncertainty quantification and point estimation, but its globally smooth latent representation is still less adaptive to localized discontinuities than the local DJGP construction. This difference is reflected in both point estimation accuracy and coverage. {The non-GP tree-ensemble baseline BART}{~attains a mid-ranked RMSE score and the second-best Cov90 rank, but one of the weakest CRPS ranks: its prediction intervals cover well without being sharp, whereas DJGP achieves comparable coverage with substantially lower CRPS.}

In summary, the synthetic experiments show that DJGP improves the overall trade-off
between point prediction and uncertainty calibration. Its advantage is most apparent when
high-dimensional inputs make vanilla local JGP unstable and when two-stage dimension
reduction loses information relevant to the downstream discontinuous surrogate. The rank
scores and coverage results together suggest that DJGP is not merely producing sharper
predictions, but is also better calibrated across heterogeneous synthetic configurations.
\endgroup

\begingroup
\subsection{Additional Robustness and Sensitivity Analyses}
\label{sec:additional_analysis}

We conducted additional experiments to examine the scaling behavior,
robustness, and tuning sensitivity of DJGP. The corresponding full
experiments are shown in Sections~D.1--D.3 of the
supplement.

\paragraph{Scaling with dimension and sample size.}
To examine whether DJGP exploits low-dimensional structure, we vary the observed
dimension $D$, intrinsic dimension $K$, and training size $N$ while holding the
other quantities fixed. RMSE remains relatively stable as $D$ increases,
deteriorates as $K$ increases, and improves with larger $N$.
These trends are consistent with Theorem~\ref{thm:6} when the approximation,
gating, and projection terms remain controlled. The oracle regression term
$\mathrm{GP}_{\mathrm{oracle}}(n_*,K)$ corresponds to a GP operating in the
$K$-dimensional latent space; under the additional oracle-design conditions in
Appendix~B.6, Corollary~2, its rate depends on $K$ and $n_*$ but not directly on
$D$. Meanwhile, larger $N$ produces denser fixed-size neighborhoods, reducing the
local-linearization term $\mathbb E[\rho_r(\V x_*)^4]$.
Thus, when the additional approximation terms remain small, predictive difficulty
is governed primarily by the intrinsic rather than ambient dimension, while denser
training data improve the local approximation. Full results are given in
Section~D.1 of the supplement.

\paragraph{Robustness to assumption violations.}
DJGP relies on two local first-order Taylor approximations: the nonlinear
dimension-reduction map is represented locally by a linear projection, and
the jump boundary is represented locally by a linear gate. We therefore
vary manifold folding and boundary curvature to probe the projection/local-
linearization terms and the gating term in Theorem~\ref{thm:6}. The
projection remains accurate under moderate locality-preserving
transformations but deteriorates once severe folding destroys latent
proximity; likewise, gate error increases when boundary curvature is large
relative to the neighborhood radius. Separately, varying jump magnitude
and observation noise tests whether these local structures remain
identifiable at a given signal-to-noise ratio (SNR). DJGP's advantage is
more evident at moderate-to-high SNR: it has the lowest fitted RMSE above
approximately \(8\) dB in this configuration and retains the
{lowest fitted CRPS} across the examined range. This advantage is most
apparent when observed neighborhoods preserve latent locality and the jump
is sufficiently distinct from observation noise for the local regime to be
identified. Section~D.2 of the supplement reports the complete stress
tests.

\paragraph{Practical tuning guidance.}
The sensitivity analyses indicate that DJGP is reasonably stable over
moderate choices of the neighborhood size, inducing sizes, and projected
dimension. Based on the configurations examined in this study, we recommend
\[
n\in[15,35],\qquad
L_1=10,\qquad
L_2=150,\qquad
Q=5,\qquad
S=300
\]
as practical starting values. Smaller neighborhoods may be preferable for
highly curved boundaries, while larger \(L_2\) can improve the projection
approximation at additional computational cost. We recommend refining
these starting values using cross-validation or a held-out validation set.
Section~D.3 of the supplement reports the full sensitivity results for
\(n\), \(L_1\), \(L_2\), \(Q\), and \(S\).
\endgroup

\begingroup
\subsection{Computation Time}
\label{subsec:computation_time}

Table~\ref{tab:runtime_method_comparison} reports wall-clock
times under the common high-dimensional LH setting. DJGP is more expensive
than the inductive baselines NGBoost and BART because it fits
target-specific local models, but it is faster than JGP-PCA and DGP in
this experiment. Section~D.4 of the supplement provides the full complexity
decomposition and empirical scaling results with respect to \(J\), \(L_1\),
and \(L_2\).

\begin{table}[t]
\centering
\small
\caption{Wall-clock time for a single model-training run on the LH benchmark
with $D=50$, $N=5000$, and $J=500$ test points.
For DJGP, the reported runtime includes PLS initialization
and subsequent model fitting with a fixed configuration,
but excludes the additional computational cost of evaluating
multiple initialization and model selection procedures used in the main
benchmark experiments.
}
\label{tab:runtime_method_comparison}
\begin{tabular}{lccccc}
\toprule
Method & NGBoost & BART & DJGP & JGP-PCA & DGP \\
\midrule
Time (s) & \textbf{10.0} & 29.5 & 36.9 & 70.5 & 93.7 \\
\bottomrule
\end{tabular}
\end{table}
\endgroup

\section{Real Dataset Experiments}\label{sec:real}
We evaluate DJGP and baseline models on three benchmark datasets from the University of California, Irvine (UCI) Machine Learning Repository{}: Wine Quality, Parkinson’s Telemonitoring, and Appliances Energy Prediction. These datasets were chosen to span a range of sample sizes, input dimensions, and, importantly, response-surface roughness {(i.e.\ the degree of local response irregularity)}. Table~\ref{tab:dataset_stats} summarizes their main characteristics, including training size $N$, input dimension $D$, test size $J$, and the latent dimension $Q$ used in the model.

To characterize local response complexity, we report three summary measures: the average local gradient $G_a$, the maximum local gradient $G_m$, and a projection-based roughness measure $\mathrm{TV}_2$. These quantities summarize average local variation, worst-case local irregularity, and overall response roughness of the regression surface, respectively.

{
For each dataset, we generate random train--test splits with sample sizes \(1000/200\)
and \(5000/500\), corresponding to the numbers of training and test observations,
respectively. We use random train--test splits for all datasets except Parkinson's
Telemonitoring, for which subject-grouped splits are used to prevent leakage across
recordings from the same patient. For each dataset setting, we repeat the experiment 20 times to account for randomness and report averaged results. Hyperparameters for all methods are selected by 5-fold cross-validation on
the first random seed, and the selected settings are then held fixed for the remaining
seeds.
}

\begin{table}[htbp]
    \centering
    \begin{threeparttable}
    \caption{Dataset statistics and smoothness metrics. {$N$ is the total size of the available dataset; the training and test subsets used in the experiments are described in the text.}}
    \label{tab:dataset_stats}
    \begin{tabular}{lrrrrrr}
        \hline
        Dataset & $N$ & $D$ & $Q$ & $G_a$ & $G_m$ & $\mathrm{TV}_2$ \\
        \hline
        Wine Quality                   & 6,497  & 11 & 3 & 0.327 & 9.90   & 8,798     \\
        Parkinson’s Telemonitoring     & 5,875  & 18 & 5 & 2.685 & 56.83  & 60,872    \\
        Appliances Energy Prediction   & 19,735 & 28 & 5 & 6.459 & 355.17 & 2,849,220 \\
        \hline
    \end{tabular}
    \begin{tablenotes}[flushleft]
        \footnotesize
        \item[1.] $G_a=\frac{1}{|\mathcal E|}\sum_{(i,j)\in\mathcal E}\frac{|y_i-y_j|}{\|\V x_i-\V x_j\|}$:
        average local gradient magnitude over the edge set $\mathcal E$ of a $k$-nearest neighbor graph ($k=6$) constructed on the training inputs.

        \item[2.] $G_m=\max_{(i,j)\in\mathcal E}\frac{|y_i-y_j|}{\|\V x_i-\V x_j\|}$:
        maximum local gradient magnitude over the same graph, summarizing worst-case local irregularity.

        \item[3.] $\mathrm{TV}_2=\sum_{i=2}^{N-1}\left|(y_{(i+1)}-y_{(i)})-(y_{(i)}-y_{(i-1)})\right|$:
        projection-based roughness measure computed after projecting the inputs onto the first principal component and ordering the responses accordingly.
    \end{tablenotes}
    \end{threeparttable}
\end{table}

\begin{table*}[ht]
\centering
\footnotesize
\setlength{\tabcolsep}{3.0pt}
\renewcommand{\arraystretch}{0.85}
\caption{UCI benchmark results under two train/test sizes.}
\label{tab:uci_all_methods}
\begin{threeparttable}
\resizebox{\textwidth}{!}{%
\begin{tabular}{llccc|ccc}
\toprule
\multirow{2}{*}{Dataset}
& \multirow{2}{*}{Method}
& \multicolumn{3}{c|}{Train=1000, Test=200}
& \multicolumn{3}{c}{Train=5000, Test=500} \\
\cmidrule(lr){3-5}\cmidrule(lr){6-8}
& & RMSE$\downarrow$ & CRPS$\downarrow$ & Cov90$\rightarrow 0.90$
  & RMSE$\downarrow$ & CRPS$\downarrow$ & Cov90$\rightarrow 0.90$ \\
\midrule

\multirow{7}{*}{Wine Quality}
 & \textbf{DJGP}
   & \textbf{0.69} & \textbf{0.39} & \textbf{0.91}
   & \textbf{0.64} & \textbf{0.34} & \textbf{0.90} \\

 & JGP
   & 0.85 & 0.51 & 0.57
   & 0.75 & 0.39 & 0.69 \\

 & JGP-PCA
   & 0.82 & 0.47 & 0.77
   & 0.80 & 0.46 & 0.74 \\

 & JGP-SIR
   & 0.80 & 0.46 & 0.78
   & 0.78 & 0.45 & 0.77 \\

 & DGP
   & 0.70 & \textbf{0.39} & \textbf{0.91}
   & 0.69 & 0.38 & 0.89 \\

 & NGBoost
   & 0.71 & 0.40 & 0.85
   & 0.70 & 0.39 & 0.89 \\

 & BART
   & 0.71 & 0.40 & \textbf{0.91}
   & 0.72 & 0.40 & 0.91 \\

\midrule

\multirow{7}{*}{{Parkinson's} Telemonitoring}
 & \textbf{DJGP}
   & \textbf{8.10} & \textbf{4.79} & \textbf{0.93}
   & \textbf{8.64} & \textbf{5.13} & \textbf{0.88} \\

 & JGP
   & 10.86 & 7.69 & 0.32
   & 10.88 & 7.78 & 0.32 \\

 & JGP-PCA
   & 10.78 & 7.56 & 0.37
   & 11.54 & 8.51 & 0.31 \\

 & JGP-SIR
   & 10.27 & 7.15 & 0.38
   & 10.22 & 7.54 & 0.30 \\

 & DGP
   & 9.24 & 6.46 & 0.32
   & 9.41 & 7.34 & 0.10 \\

 & NGBoost
   & 9.65 & 6.68 & 0.41
   & 9.34 & 6.50 & 0.39 \\

 & BART
   & 9.09 & 5.59 & 0.71
   & 8.66 & 5.32 & 0.73 \\

\midrule

\multirow{7}{*}{Appliances Energy Prediction}
 & \textbf{DJGP}
   & 101.26 & \textbf{40.87} & \textbf{0.90}
   & \textbf{97.74} & \textbf{37.19} & \textbf{0.92} \\

 & JGP
   & 138.30 & 63.51 & 0.45
   & 118.48 & 48.52 & 0.48 \\

 & JGP-PCA
   & 119.32 & 52.29 & 0.62
   & 108.16 & 42.74 & 0.65 \\

 & JGP-SIR
   & 111.38 & 47.27 & 0.64
   & 116.45 & 44.31 & 0.64 \\

 & DGP
   & 101.10 & 47.72 & 0.92
   & 99.00 & 46.00 & 0.93 \\

 & NGBoost
   & \textbf{97.99} & 42.16 & 0.86
   & 97.80 & 39.97 & \textbf{0.92} \\

 & BART
   & 99.56 & 46.38 & 0.93
   & 102.40 & 47.10 & 0.93 \\

\bottomrule
\end{tabular}%
}
\vspace{2mm}
\begin{minipage}{0.96\textwidth}
\footnotesize
\emph{Note.} RMSE and CRPS are better when smaller. Cov90 is better when closer
to the nominal 0.90 level, not necessarily when larger. Bold entries indicate
the best or rounded-tied best method for each metric within each dataset and train/test size
according to this rule.
\end{minipage}
\end{threeparttable}
\end{table*}

\begingroup
The UCI benchmark results in Table~\ref{tab:uci_all_methods} are broadly consistent with
the trends observed in the synthetic experiments, while also illustrating the behavior of
DJGP on general tabular regression tasks that do not explicitly contain jump
discontinuities. Overall, DJGP remains competitive with strong probabilistic baselines
across the three datasets and two train/test sizes. Its advantage is most pronounced in
CRPS and Cov90, suggesting that the local projection model provides a favorable
balance between point accuracy and predictive calibration.

The two-stage JGP variants with external dimension reduction, JGP-PCA and JGP-SIR, are
typically weaker than the global probabilistic baselines NGBoost and DGP. This is
consistent with the synthetic results: applying dimension reduction as a preprocessing step
can remove information that is useful for the downstream local probabilistic model,
especially when the projection is not learned jointly with the local likelihood. In contrast,
DJGP learns the low-dimensional projection jointly with the local GP/JGP component, which
helps avoid the loss of predictive information introduced by fixed preprocessing.

NGBoost and DGP are strong baselines and provide useful points of comparison.
NGBoost is often competitive in RMSE, reflecting the strength of boosted tree-based
probabilistic regression for tabular data. However, its Cov90 values are sometimes below
the nominal level, indicating a tendency toward overconfident predictive intervals. DGP is competitive in several cases and can provide
reasonable coverage, but its global smooth latent representation does not always improve
over simpler probabilistic alternatives on these tabular datasets.

In contrast to the synthetic jump tasks, BART is considerably more competitive on these real benchmarks, ranking as the strongest non-DJGP method on several cells. This is most pronounced on Parkinson's Telemonitoring, where BART attains the second-best RMSE and CRPS after DJGP---nearly matching DJGP's RMSE on the larger split---and substantially better coverage than the GP baselines, although it still under-covers relative to DJGP.

The Parkinson's Telemonitoring dataset deserves separate interpretation. We use grouped
train/test splits so that observations from the same patient do not appear in both training
and test sets. This creates a more difficult extrapolation problem across subjects rather
than a standard random interpolation task. Under this grouped split, the larger
\(5000/500\) setting is not necessarily easier: the larger test set can include more
out-of-distribution patient-level variation, and several flexible models, including DJGP and
DGP, show some degradation relative to the smaller split. DJGP nevertheless maintains the
best or near-best uncertainty calibration on this dataset, which is important for
patient-level extrapolation.

In summary, although the UCI datasets are not designed to exhibit explicit jump behavior,
DJGP remains competitive with strong non-GP and deep-GP baselines on general tabular
regression. These results suggest that the proposed local projection model is useful beyond
data with jump behavior, providing competitive probabilistic predictions in general surrogate
modeling problems as well, particularly when calibrated uncertainty is important.
\endgroup

\section{Conclusion}\label{sec:conclusion}
We introduced the Deep Jump Gaussian Process, a surrogate model that
{combines globally coupled local projection learning with local discontinuity
detection}. By placing Gaussian process priors on target-specific projection matrices and integrating dimension reduction into JGP, DJGP jointly learns low-dimensional representations and piecewise-continuous regimes in high-dimensional inputs. A gradient-based variational inference algorithm enables joint optimization of projection parameters, local JGP hyperparameters, and partitioning structure, while inducing-point approximations ensure scalability.

On the theoretical side, we derived a finite-sample risk decomposition that
separates DJGP prediction error into approximation, gating, projection,
local-linearization, and latent-space oracle-regression components.

Across synthetic benchmarks, DJGP achieves the best aggregate rank {on RMSE, CRPS, and coverage}; on three real-world UCI datasets it remains competitive in RMSE and CRPS while retaining strong calibration. Its integrated dimensionality reduction mitigates overfitting in local neighborhoods and improves predictive reliability in sparse, high-dimensional settings.

DJGP’s ability to capture abrupt regime changes while providing uncertainty quantification makes it well suited for applications such as materials science, econometrics, and social sciences, where structural shifts are common.

\begingroup
\paragraph{Limitations.} DJGP relies on locality-preserving observed neighborhoods and locally
approximately linear gates. When the observation map severely folds the latent manifold or the
boundary curvature is high relative to the neighborhood radius, performance can degrade; our
assumption-check experiment shows that the projection gap grows sharply under severe folding. The
transductive construction also makes computation scale with the number of prediction anchors.
DJGP can also lose its point-accuracy advantage when the jump signal is weak relative
to observation noise, because a small local neighborhood may then contain insufficient
information to distinguish a genuine regime change from noise;
{Section~D.2 of the supplement quantifies this low-signal-to-noise regime.}
Finally, reliable early stopping and hyperparameter selection remain practical challenges.
\endgroup

\section*{Data Availability Statement} 
The Python implementation of DJGP, together with a lightweight end-to-end reproducibility benchmark, is available at \url{https://github.com/crushonyfg/DJGP_pub}. The released benchmark uses fixed subject-disjoint Parkinson's Telemonitoring splits for seeds 0, 1, and 2 and can be reproduced by running \texttt{python reproduce\_parkinson.py}. The script generates both per-seed and aggregated evaluation results, which can be directly compared with the values reported in Section~D.7 of the supplement. The complete experimental protocols for the main synthetic and real-data studies are described in Sections~\ref{sec:simulations}--\ref{sec:real} and the supplement.

\if0\blind
{
\section*{Acknowledgment}
We acknowledge support for this work from the Air Force Office of Scientific Research (FA9550-23-1-0673) and the National Science Foundation (NSF-2420358).
} \fi

\section*{Conflict of Interest Statement}
The authors report there are no competing interests to declare.

\spacingset{1.0}
\bibliographystyle{agsm}
\begingroup
\small
\bibliography{djgpbib}
\endgroup

\end{document}